%% file: paper.tex
\documentclass[conference]{ieeeconf}	% A4 paper: add option [a4paper]

\IEEEoverridecommandlockouts			% This command is only needed if you want to use the \thanks command

\usepackage{censor}
\StopCensoring

\makeatletter
\let\NAT@parse\undefined
\makeatother
\usepackage[hidelinks]{hyperref}

\usepackage{cite}

\usepackage{amssymb}
\usepackage{amsmath}

\usepackage{multirow}

\usepackage{flushend}

\usepackage{algorithm}
\usepackage{algpseudocode}

\algrenewcommand{\algorithmicensure}{\textbf{Given:}}
\algrenewcommand{\algorithmicrequire}{\textbf{Input:}}

\algrenewcommand{\algorithmiccomment}[1]{\hfill {\color{gray} $\triangleright$ #1}}	%\small 

\makeatletter
\renewcommand{\ALG@beginalgorithmic}{\footnotesize}
\makeatother

\usepackage[dvipsnames]{xcolor}
\usepackage{svg}

\input{tikz_styles}

\newcommand{\eg}{e.\,g.,~}

\newcommand{\shot}{TrySee}
\newcommand{\chaser}{Modular Perception-Enhanced VLM Grounding for\\Closed-Loop Robot Task Planning in Specialized Domains}
\title{\LARGE \bf
	\shot: \chaser
} 

\author{
	\censor{Elias Hofele}$^{1,2}$,~
	\censor{Thomas Trapp}$^1$,~
	\censor{David Timmermann}$^1$,\\
	\censor{Tristan Schnell}$^1$,~
	\censor{Georg Heppner}$^1$,~
	\censor{Arne Roennau}$^{1,2}$
	\thanks{$^1$\xblackout{FZI Research Center for Information Technology
		\texttt{\string{lastname\string}@fzi.de}}
	}%
	\thanks{$^2$\xblackout{Machine Intelligence and Robotics Lab (MaiRo),
		Karlsruhe Institute for Technology (KIT)}
	}%
	\thanks{The research leading to this work has received funding in 
		\xblackout{the \mbox{GANResilRob} project under Grant Agreement No. 01MJ22003A by the Federal Ministry for Economic Affairs and Energy (BMWE) on the basis of a decision by the German Bundestag,
		and the AIDeCo project, which is supported by donations from the Intel Corporation.}
	}%
}%

\begin{document}
	\maketitle

	\input{sections/abstract}

	\input{sections/introduction}

	\input{sections/sota}

	\input{sections/approach}

	\input{sections/experiments}

	\input{sections/results}

	\input{sections/conclusion}

	\input{sections/acknowledgments}

	\bibliographystyle{IEEEtran}
	\bibliography{IEEEabrv, paper}

\end{document}

%% file: tikz_styles.tex
\usepackage{tikz}
\usetikzlibrary{math, calc, arrows.meta, decorations.text, bending}

\tikzset{
  setup/.style={
    every node/.style={
      inner sep=0,
      outer sep=0,
      anchor=center
    }
  },
  pipeline/.style={
    /utils/exec={
      \sffamily\fontsize{6}{7pt}\selectfont
    },
    every node/.style={
      inner sep=0,
      outer sep=0,
      anchor=center
    },
    block/.style={
      rectangle,
      rounded corners
    },
    component/.style={
      block, 
      align=center,
      font=\normalsize
    },
    vlm/.style={
      component,
      minimum width=2.3cm,
      text width=2.1cm, 
      minimum height=1.4cm,
    },
    data/.style={
      block,
      align=left,
      anchor=north west,
      fill=Gray!70
    },
    planning output/.style={
      fill=Goldenrod!40
    },
    checking output/.style={
      fill=SeaGreen!30
    },
    vision output/.style={
      fill=CornflowerBlue!30
    },
    photo/.style={
      anchor=north west,
      clip,
      rounded corners
    },
    group/.style={
      block,
      dashed,
      draw=darkgray, 
      anchor=west
    },
    label/.style={
      anchor=south west,
      yshift=-1pt,
      text=darkgray
    },
    number/.style={
      circle,
      draw=darkgray,
      fill=darkgray,
      anchor=center,
      text=white,
      minimum width=.3cm
    },
    arrow base/.style={
      draw=Gray!50,
      line width=2mm,
      rounded corners=4mm,
      shorten >=1mm,
    },
    arrow/.style={
      arrow base,
      -{Triangle[angle=30:1pt 3, length=3mm, flex]}
    },
    double arrow/.style={
      arrow base,
      shorten <=1mm,
      {Triangle[angle=30:1pt 3, length=3mm]}-{Triangle[angle=30:1pt 3, length=3mm]}
    },
    arrow label/.style={
      align=center,
      midway,
      sloped
    },
  },
  collage/.style={
    every node/.style={
      inner sep=0,
      outer sep=0
    },
    image label/.style={
      fill=black,
      text=white,
      anchor=north west,
      xshift=0.1cm,
      yshift=-0.1cm,
      inner sep=0.02cm,
      font=\footnotesize
    },
    image/.style={
      anchor=north west
    },
  }
}

%% file: sections/abstract.tex
\begin{abstract}
    Vision Language Models (VLMs) have shown impressive capabilities in high-level reasoning and task planning for robots.
    However, specialized domains remain a challenge for VLMs' image interpretation and grounding abilities.
    We introduce \shot, a modular closed-loop robot task planning pipeline integrating general-purpose VLM reasoning with domain-specific perception and established robotics components.
    The system consists of an inherently reactive \textit{cyclic planning pipeline}, augmented with domain-specific \textit{modular vision preprocessing}, to provide online grounding.
    Finally, a \textit{dynamic preprocessing selection} mechanism contributes to the scalability of the pipeline. 
    The system requires neither VLM fine-tuning nor end-to-end demonstrations and adapts through replaceable perception modules and motion planning.
    The approach is evaluated on planning tasks across 5 widely different domains, including a real-world robot disassembly task.
    The field tests in a brownfield environment demonstrate the approach's practicality in integrating VLM planning into specialized robotic systems with limited domain-specific training data.
\end{abstract}

% \begin{keywords}
%     Semantic Grounding, Vision Language Models, Robot Task Planning, Robot Perception
% \end{keywords}

%% file: sections/introduction.tex
\section{Introduction}
\label{sec:intro}

The idea of controlling robots with natural language has long been a staple of science fiction.
However, accomplishing even simple tasks, such as \textit{``Clean up the table.''}, requires a complex understanding of the world and the robot's capabilities, as well as drawing on many of the core challenges in robotics, such as perception, manipulation, and planning.

With the recent advances in Vision Language Models (VLMs)~\cite{liu2023visualinstructiontuning}~\cite{zhu2023minigpt4enhancingvisionlanguageunderstanding}, significant progress has been made in enabling robot reasoning and planning by leveraging vast amounts of real-world knowledge~\cite{ahnCanNotSay2022}~\cite{rana2023sayplan}.
While VLM-based systems have produced impressive results for tasks in everyday environments (such as offices or homes), they still struggle in more specialized domains, including industrial and scientific applications~\cite{gu2024anomalygpt}~\cite{zhang2024vision}.
This is attributable to these domains being underrepresented in the VLMs' training data. 
The resulting performance gap is a barrier to the more widespread adoption of VLMs in these fields, in which data collection and annotation is often particularly expensive and time-consuming.

\begin{figure}[t]
    \input{figures/hero}
    \caption{Example images used to evaluate \shot, preprocessed by the pipeline's Vision Preprocessing Module.
    The images are labeled A through E, corresponding to the tasks used to evaluate the proposed pipeline (see Table~\ref{tab:tasks}).
    \textbf{A} shows a computer assembly task, \textbf{B} an agricultural task, \textbf{C} a service task, \textbf{D} an aircraft cockpit interaction task, and \textbf{E} an e-waste disassembly task.
    Additional examples are collected in Fig.~\ref{fig:collage}.}
    \label{fig:hero}
\end{figure}
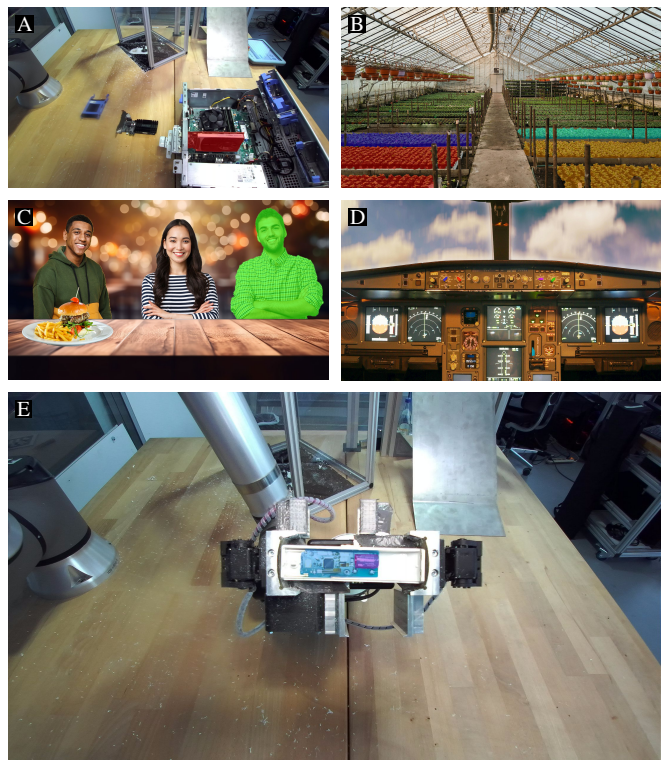

A fundamental requirement for VLMs to generate effective robot task plans is the ability to ground their language understanding in the robot's perception and action capabilities.
In other words, the objects and features comprising the robot's environment must be linked to the semantics used in the given task, and these, in turn, must be linked to the concrete movements and poses necessary to interact with them.
General-purpose VLMs often fail at the first step for recondite objects, and their image-level outputs alone do not provide sufficiently reliable geometric targets for robot execution.
A practical planning system therefore requires explicit interfaces connecting domain-specific perception, VLM-level semantics, and the robot's action capabilities.

In this work, we tackle the challenge of VLM grounding in specialized domains by presenting a modular pipeline for VLM-based robot task planning.
Rather than fine-tuning the model or replacing the entire existing robot software stack, \shot~compensates the  general-purpose VLM's limitations in object recognition and robot pose generation by connecting its reasoning abilities with replaceable domain-specific perception modules and established motion-planning components.
This allows the pipeline to be adaptable to different domains and tasks, through targeted perception models without requiring end-to-end robot demonstrations.

In \shot, the VLM incrementally generates and verifies short-horizon action sequences, while the perception modules provide semantic object labels, visual masks, and executable target poses.
This provides online grounding that is continuously updated as the environment changes.
For object-dense environments, the VLM can request preprocessing only for the information relevant to the current verification step.

Our approach enables the VLM to generate task plans for a variety of domains.
We evaluate the pipeline using tasks across 5 vastly different domains (see Fig.~\ref{fig:hero}), including e-waste recycling, a workshop environment, and an agricultural setting.
The focus is on complex, multi-step tasks.
In order to underline the real-world applicability of our approach, we also implement the pipeline on a real robot and evaluate it in an industrial recycling environment.

%% file: figures/hero.tex
%\tikzsetfigurename{hero}
\begin{tikzpicture}[collage]
    \tikzmath{
        \imagewidth = .49\linewidth;
        \imagespacing = .02\linewidth;
    }
    \node[image] (A) at (0, 0) {
        \includegraphics[width=\imagewidth pt]{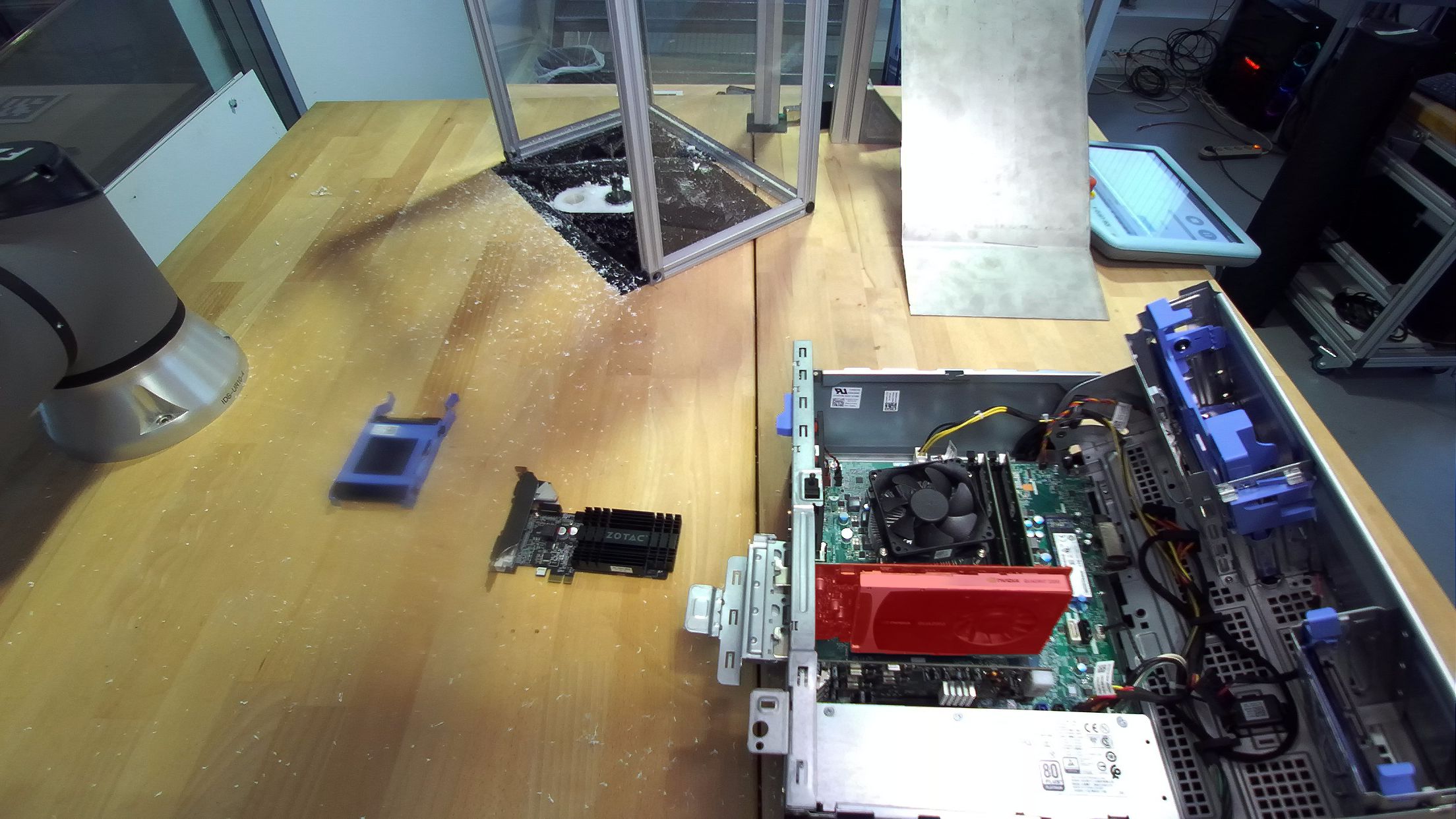}
    };
    \node[image label] at (A.north west) {A};

    \node[image] (B) at ($(A.north east) + (\imagespacing pt, 0)$) {
        \includegraphics[width=\imagewidth pt]{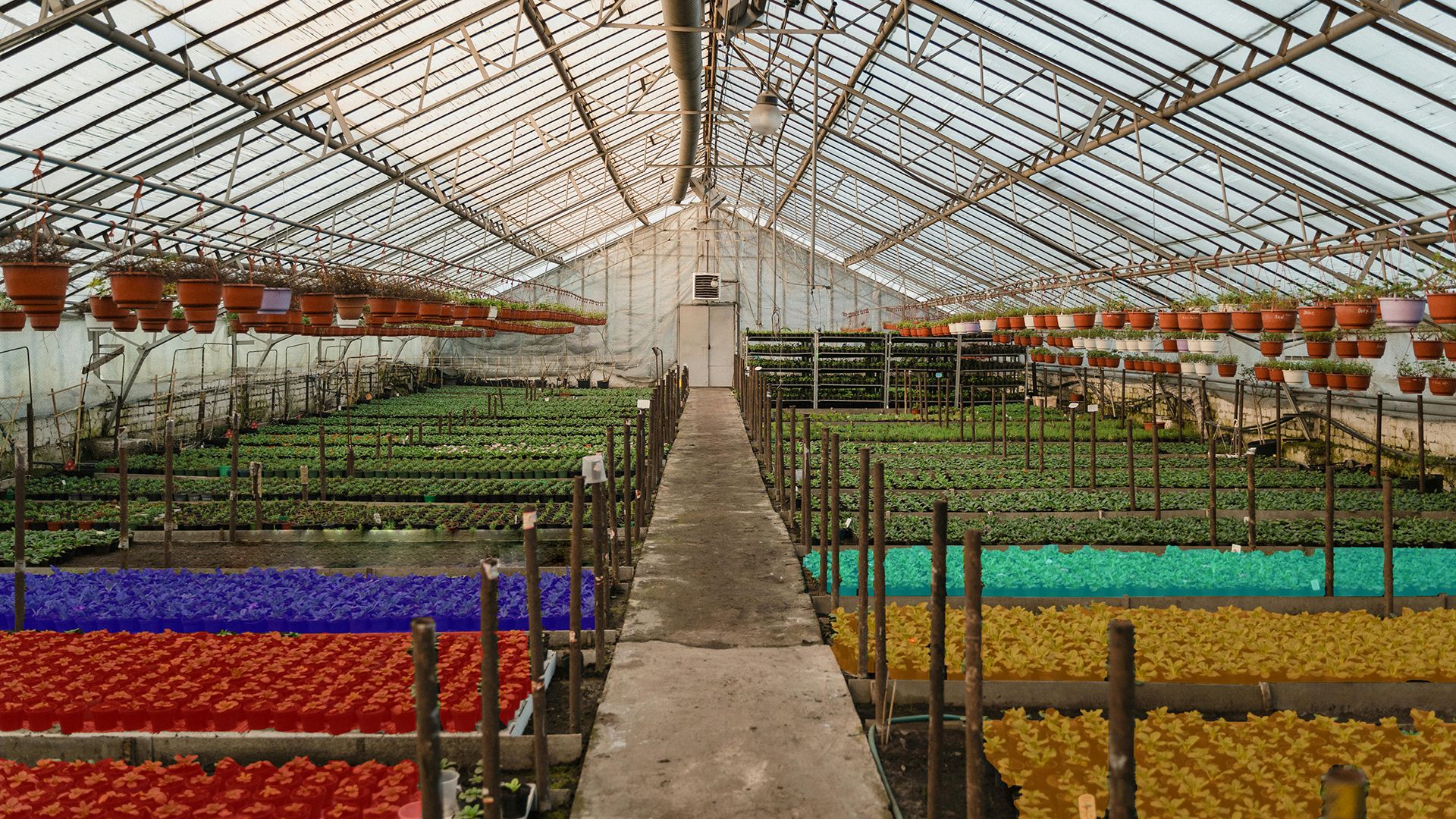}
    };
    \node[image label] at (B.north west) {B};
    \node[image] (C) at ($(A.south west) + (0, -\imagespacing pt)$) {
        \includegraphics[width=\imagewidth pt]{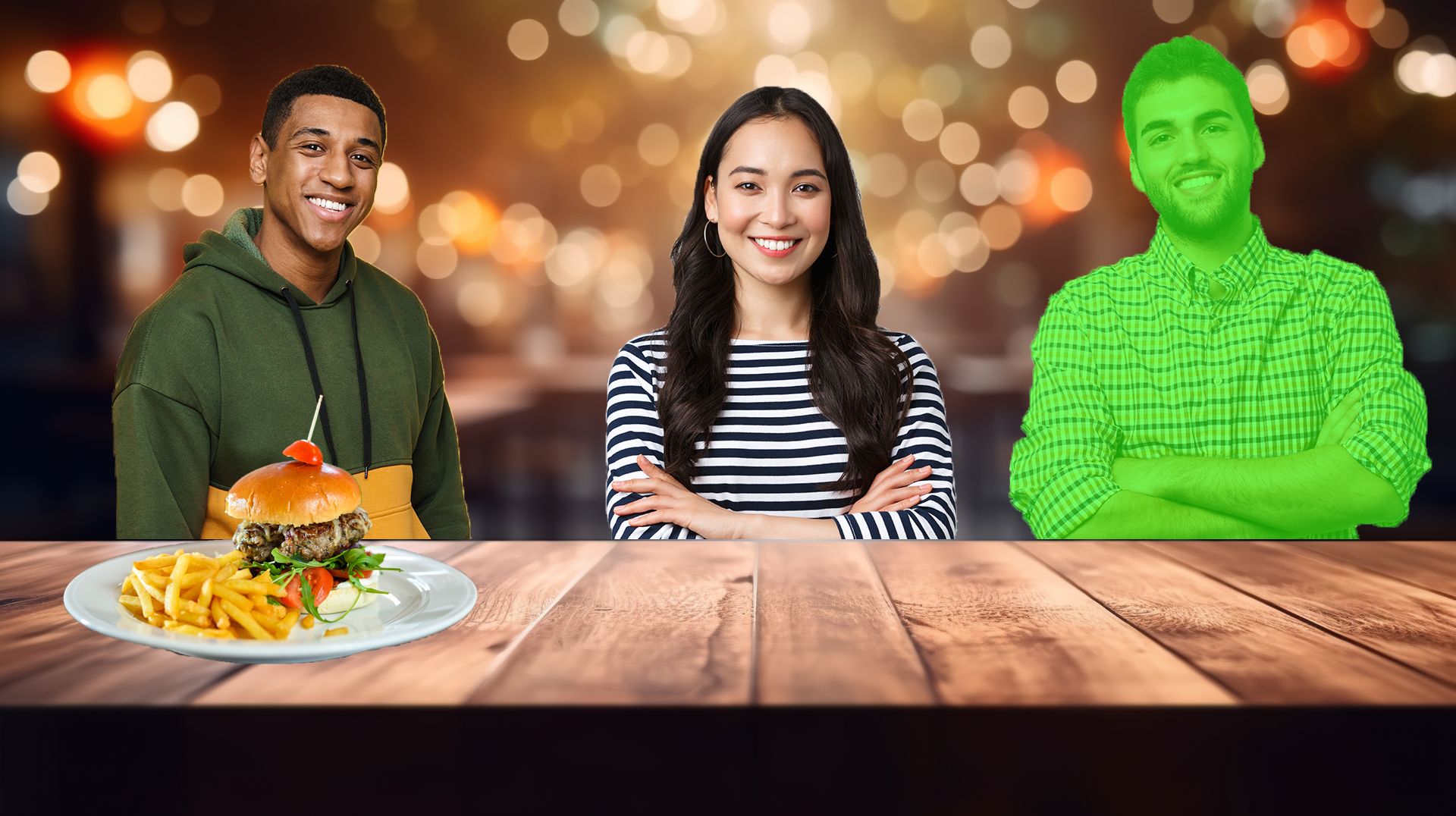}
    };
    \node[image label] at (C.north west) {C};

    \node[image] (D) at ($(C.north east) + (\imagespacing pt, 0)$) {
        \includegraphics[width=\imagewidth pt]{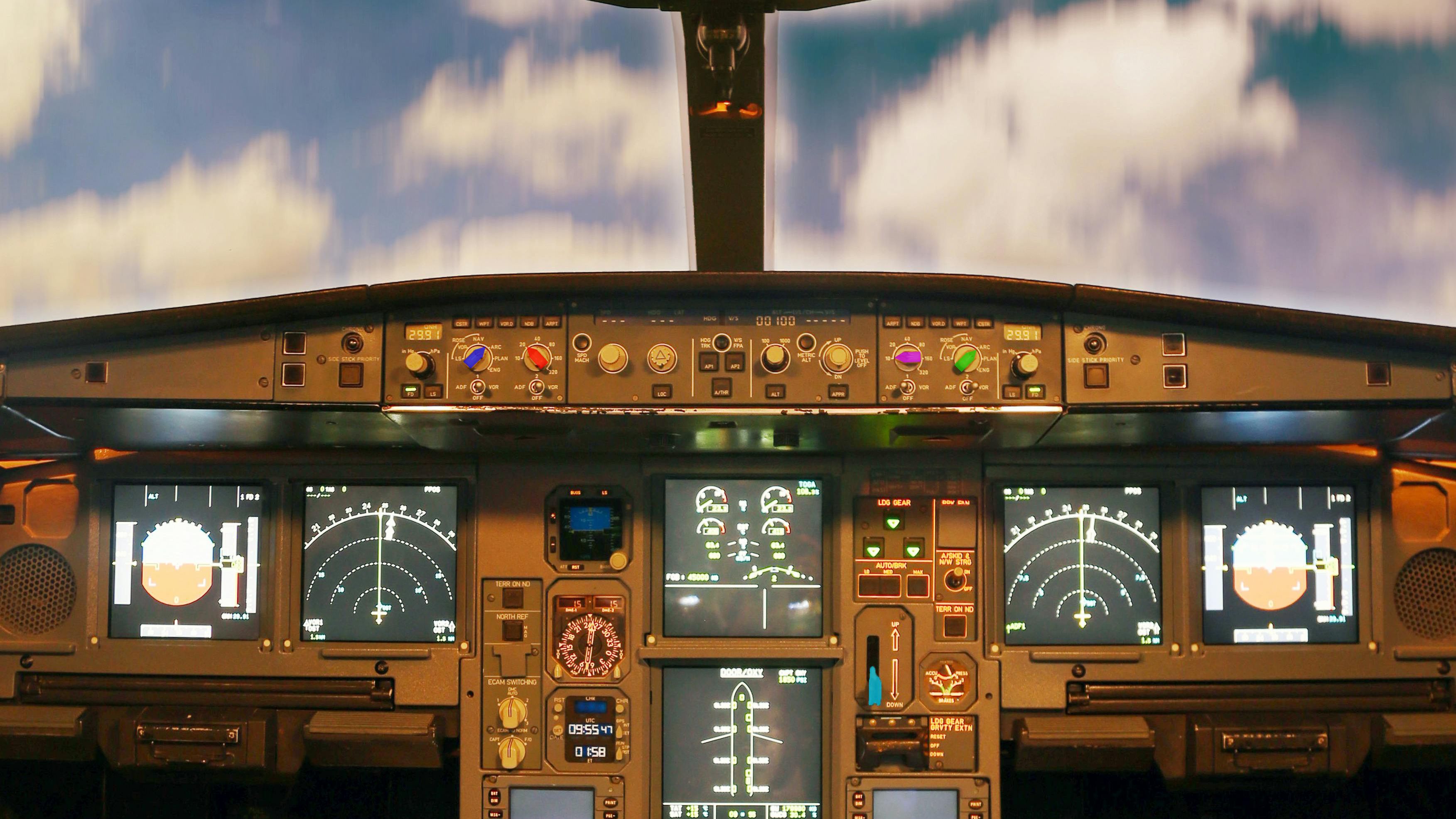}
    };
    \node[image label] at (D.north west) {D};
    \node[image] (E) at ($(C.south west) + (0, -\imagespacing pt)$) {
        \includegraphics[width=\linewidth]{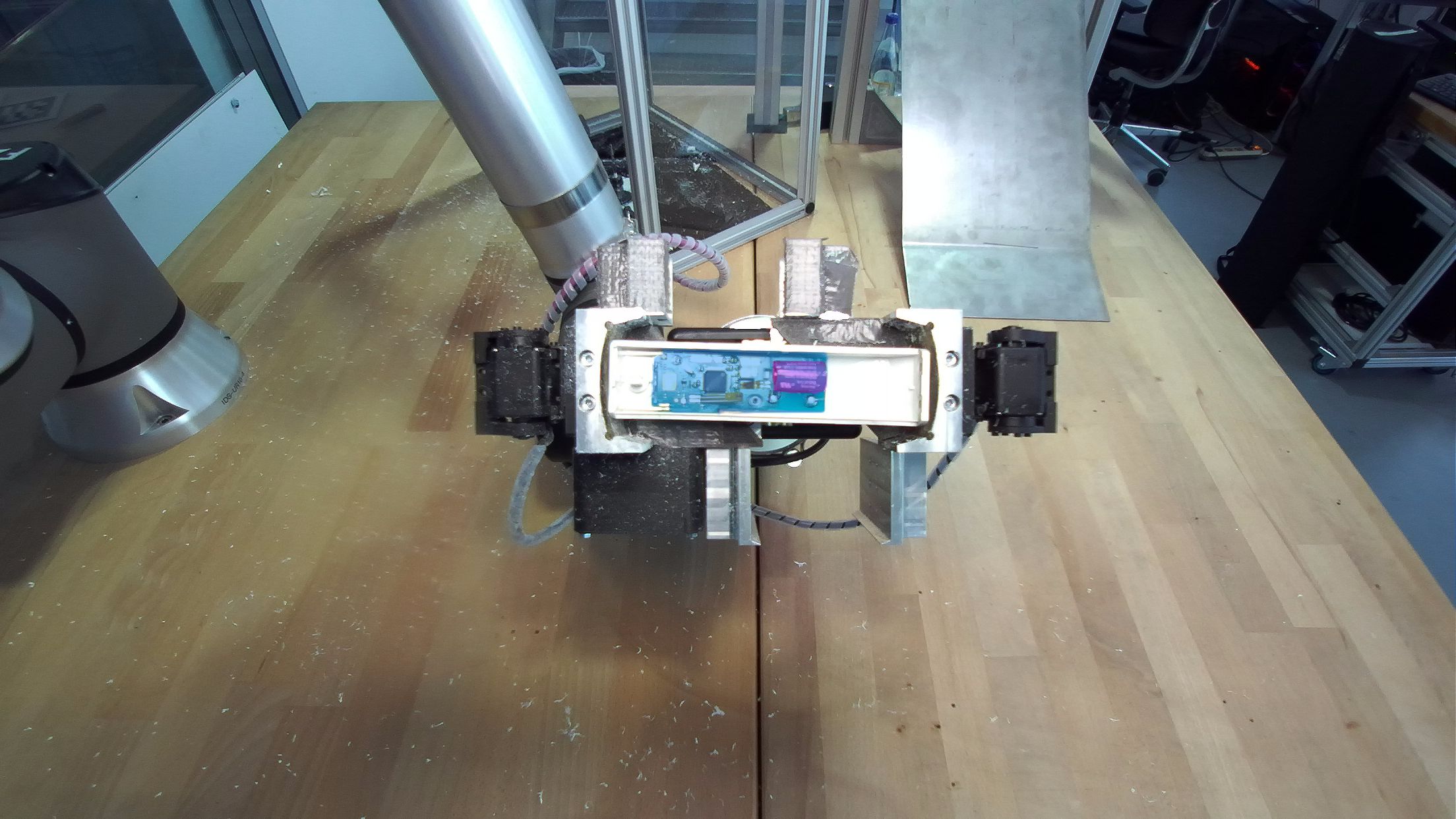}
    };
    \node[image label] at (E.north west) {E};
\end{tikzpicture}

%% file: sections/sota.tex
\section{Related Work}
\label{sec:sota}

In recent years, substantial progress has been made in the application of Large Language Models (LLMs) and Vision Language Models (VLMs) in high-level reasoning and task planning for robots.

\textbf{LLM- and VLM-based task planning:}
In \texttt{SAYCAN}~\cite{ahnCanNotSay2022}, planning is performed by sequentially selecting skills from a predefined library.
Candidate skills are scored using both their linguistic relevance and learned affordance functions, grounding their selection in the robot's capabilities.
\texttt{PROGPROMPT}~\cite{singhProgPromptGeneratingSituated2022} leverages the code-generation capabilities of modern LLMs by expressing actions, assertions, and plans directly in Python syntax, making them easier to process downstream. 
\texttt{SAYPLAN}~\cite{rana2023sayplan} grounds planning in large environments by traversing a 3D scene graph representation. 

These approaches represent open-loop task planners and are easily prone to failure in dynamic environments. 
By contrast, our approach and the remainder of this section discuss closed-loop task planning approaches instead. 

\textbf{Reactive task planning using LLMs and VLMs:}
\texttt{INNERMONOLOGUE}~\cite{huangInnerMonologueEmbodied2022} attempts to close the loop of \texttt{SAYCAN} by injecting execution feedback and scene descriptions from visual-question-answering models (or human feedback) into the LLM prompt at each step.
\texttt{ROSLLM}~\cite{mowerROSLLMROSFramework2024} implements a generic ROS-based framework for integrating LLM agents with robotic systems and supports task correction based on feedback cues.
\texttt{REFLECT}~\cite{liuREFLECTSummarizingRobot2023} maintains a hierarchical memory structure and semantic scene representation, extracted via object detection and CLIP to find relations, to support failure analysis and replanning. 
However, the LLM itself is only fed textual prompts. 
\texttt{BETR-XP-LLM}~\cite{styrudAutomaticBehaviorTree2024} and \texttt{REPLAN}~\cite{meiReplanVLMReplanningRobotic2024} employ VLM-based verification and critique mechanisms to detect and correct planning failures. 
\texttt{AHA}~\cite{duanAHAVisionLanguageModelDetecting2024} further studies failure detection through task-specific fine-tuning of VLMs.
More recent work investigates VLM-based failure detection and recovery through optimized prompting~\cite{chenAutomatingRobotFailure2024}, behavior-tree-based reactive failure handling~\cite{ahmadUnifiedFrameworkRealtime2025}, and constraint-aware visual monitoring~\cite{zhouCodeMonitorConstraintaware2025}. 
Other approaches represent plans as behavior trees~\cite{colledanchise_behavior_2022}, allowing the resulting policies to be executed directly within established robotic control frameworks~\cite{izzoBTGenBotBehaviorTree2024,zhouLLMBTPerformingRobotic2024,lykovLLMBRAInAIdrivenFast2023,lykovLLMMARSLargeLanguage2023,chenIntegratingIntentUnderstanding2024}.

These works demonstrate the importance of reactive execution. 
However, the above approaches rely on external grounding mechanisms to execute the generated plans.
Conversely, in our approach the grounding is performed online through the vision module itself.

\textbf{Explicit visual grounding:}
\texttt{ConceptGraphs}~\cite{guConceptGraphsOpenVocabulary2023} uses 3D scene graph representations to support language-conditioned planning and spatial reasoning. 
Other work explores multimodal interfaces between vision-language models and symbolic planners~\cite{shiraiVisionLanguageInterpreterRobot2023}. 
Recent segmentation-oriented VLMs such as \texttt{LISA}~\cite{laiLISAReasoningSegmentation2024} show that segmentation masks can be generated from language-based reasoning queries, highlighting the relevance of explicit visual grounding for downstream robotic tasks.
Related visual-prompting approaches explicitly annotate candidate image regions: \texttt{SoM} (Set-of-Mark Prompting) uses interpretable marks to improve visual grounding in multimodal models~\cite{yangSetOfMarkPrompting2023}.
Similarly, \texttt{MOKA} employs mark-based visual prompts and point-based affordances to connect VLM predictions to robot manipulation actions~\cite{fangMOKAOpenWorld2024}.

Our work follows a deployment-oriented philosophy by integrating modular perception components directly into the planning loop.
This allows the planning system to continuously update its grounding during execution while remaining adaptable to different domains through the replacement of individual perception modules.

\textbf{Vision Language Action (VLA) models:} More recently, VLA models have been proposed to directly generate actions from visual inputs~\cite{brohan2023rt2visionlanguageactionmodelstransfer, kim2024openvlaopensourcevisionlanguageactionmodel, nvidia2025gr00tn1openfoundation, black2026pi0visionlanguageactionflowmodel}. 
These models are trained on large datasets of teleoperation data, thus learning the relationship between visual features and actions. 

However, VLA models generally require substantial amounts of training data and are often tied to specific robot embodiments, action spaces, or domains. 
Adapting them to highly specialized applications remains challenging when representative demonstration data is unavailable. 
By contrast, our approach operates at the task-planning level and combines VLM reasoning with modular perception and classical motion generation techniques. 
This enables the reuse of existing robotic software stacks while facilitating adaptation to new domains through targeted perception modules rather than large-scale retraining.

%% file: sections/approach.tex
\begin{figure*}[!ht]
    \centering
    \vspace{1ex}
    \input{figures/pipeline/pipeline}
    \caption{\shot~is a cyclic planning pipeline, with the numbers in the top-right corners indicating the flow of operations.
    In each iteration, the next step of the overall plan is generated, executed and verified.
    The Vision Preprocessing Module is central to the pipeline, and interacts with every one of the other components.
    The images shown were not used in the experiments, but purely serve to illustrate the pipeline's operation.}
    \label{fig:pipeline}
\end{figure*}
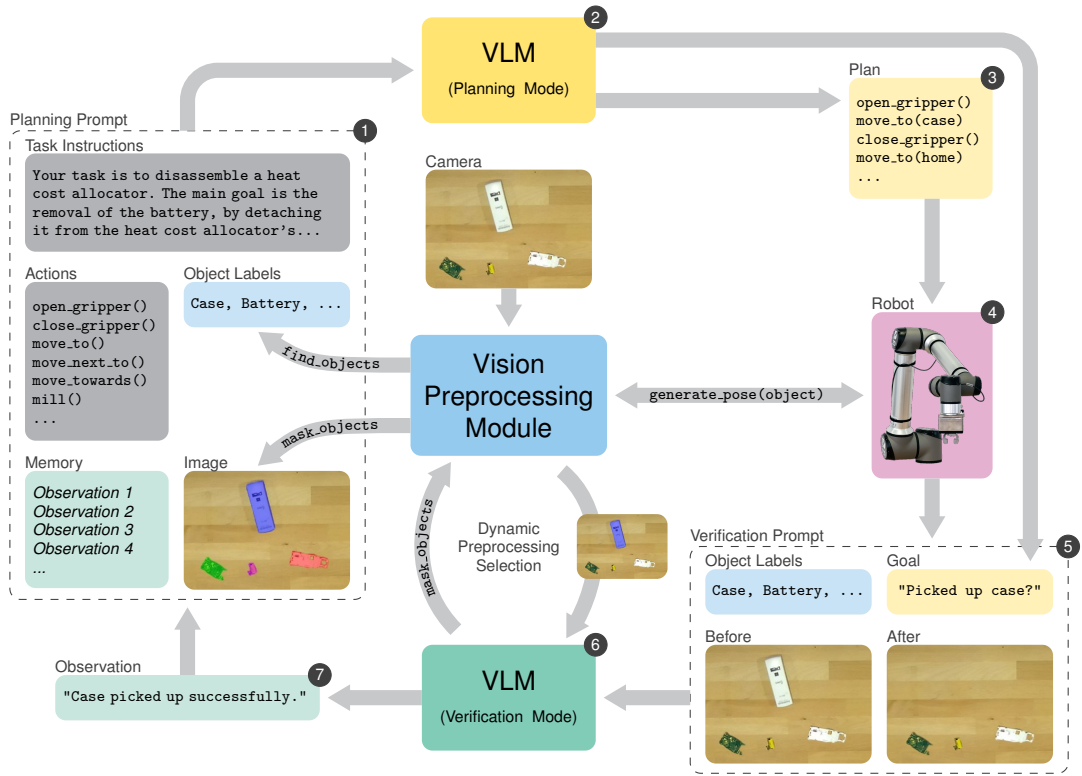

\section{\shot}
\label{sec:approach}

This work aims to improve the high-level task planning capabilities of robots for tasks from specialized domains based on instructions in natural language.
In this context, a specialized domain is defined as comprising a specific set of objects, features, or components that are not found in general-purpose datasets and thus are unlikely to be recognized by the VLM out-of-the box. 
In particular, the focus is on improving the practical applicability of VLM-based robotic planning systems.
An overview of the proposed approach is shown in Fig.~\ref{fig:pipeline}.

\subsection{Components}
\label{subsec:approach_components}
In this section, we introduce the two main components \shot~is based on: a Vision Language Model (VLM) and a Vision Preprocessing Module.
Since the focus of this work is on the integration of these components into a pipeline, we discuss the high-level functionality of the components and how they interact with each other.
For more in-depth details on one possible implementation of these components, see the Evaluation Setup in Section~\ref{sec:experiments}.
Additionally, the underlying robot execution stack is treated as existing infrastructure and exposed through the action toolbox in Table~\ref{tab:toolbox}.

\begin{table}[b]
    \centering
    \caption{Toolbox of actions available to the robot.}
    \input{figures/toolbox}
    \label{tab:toolbox}
\end{table}

\paragraph*{Vision Language Model} For this work, the VLM is employed in two different modes, differing in how they are prompted.
The alternating use of these two modes forms the basis of the \textit{cyclic planning pipeline}.
On the one hand, the VLM is used to generate the next planning step based on the task at hand and the current state of the environment.
To this end, it is provided with the task description, the list of actions the robot can perform (each of which represents a basic motion primitive, as seen in Table~\ref{tab:toolbox}, which can be combined with others to form more complex behaviors), current images of the environment, and the list of interactable objects and points of interest found therein.
In turn, the VLM generates a \textit{Thought} (a natural language description of the next step to be executed), a sequence of actions from the toolbox, as well as the goal it aims to achieve with the generated sequence (in natural language).
The \textit{Thought} provides an explicit rationale alongside the structured action output.
The division of the overall plan into smaller steps is entirely self-directed by the VLM, which is instructed to generate only one major interaction with its surroundings (\eg picking up an object) at a time.

Additionally, the same VLM is also used to verify whether the generated plan was executed successfully.
This is accomplished by comparing the goal generated in the planning step against current images of the environment as well as images from directly before the execution of the plan.
The model first describes the images provided before generating an \textit{Observation} stating whether or not the expected goal was achieved.
The collection of past \textit{Observations} comprises the memory of the VLM, which is also appended to its input in subsequent planning steps.

The division of the pipeline into explicit planning and verification steps is a key design choice.
It allows the VLM to focus on each step individually, without being overwhelmed by having to perform both of these complex tasks simultaneously.
Separate prompts are used for planning and verification so that each invocation has a constrained objective and output format.

For the purposes of the proposed pipeline, the VLM is treated as a black box, with the models not being fine-tuned or adapted to the specific task at hand.
This separates domain adaptation from the choice of reasoning model and allows newer VLMs to be integrated without modifying the perception or execution interfaces.
Additionally, while fine-tuning the VLM might improve the model's innate scene understanding capabilities, external modules would still be required to provide the necessary action grounding.

\paragraph*{Vision Preprocessing Module} The \textit{modular vision preprocessing} is provided by a collection of dedicated computer vision systems, which together aid in grounding the VLM's language understanding by providing three functions.
Despite its name, the module serves both as a visual preprocessing interface for the VLM and as the action-grounding interface for the robot execution layer.
Initially, a \texttt{find\_objects} function identifies any task-relevant interactable objects in the environment and returns their (expressive, natural language) labels.
The VLM then uses these labels to reason about the objects in the environment and to parametrize the resulting action sequence.
This creates a shared vocabulary between the VLM and the Vision Preprocessing Module.
Secondly, a \texttt{mask\_objects} function is used to overlay a segmentation mask of the given object(s) on the input image.
The colors of the masks are communicated to the VLM together with the object labels and masked image, allowing it to reason about the spatial relations of the object in the environment.
This performs the task of grounding the VLM's language understanding in the visual input.
Finally, a \texttt{generate\_pose} function is used to generate a target pose for the given object label.
Applying this function for each label used in the task plan enables the robot to actually interact with its surroundings, grounding its actions in the real world.

\begin{table*}[hbt]
    \centering
    \vspace{1.2ex}
    \caption{Overview of the tasks (with respective domains \& required object recognition capabilities) used to perform the evaluation.}
    \input{figures/tasks}
    \label{tab:tasks}
\end{table*}

\subsection{Pipeline}
\label{subsec:approach_pipeline}
Fig.~\ref{fig:pipeline} shows an overview of the pipeline.
The implementation is based on the ReAct framework~\cite{Yao2023}, with the actions from Table~\ref{tab:toolbox} being used as the tools of a VLM-based agent.

The pipeline starts by taking current images of the workspace and preprocessing them using the \texttt{find\_objects} and \texttt{mask\_objects} functions introduced above.
The resulting images are added to the VLM's planning mode input, which uses them to generate an action sequence.
The generated actions are parsed into a Behavior Tree (BT), whose leaf nodes invoke the corresponding functions from the robot's action toolbox.
The BT provides an inspectable interface between the VLM-generated plan and the underlying robot-control framework.
Any object labels parametrizing the plan are mapped to the corresponding robot pose using the Vision Preprocessing Module's \texttt{generate\_pose} function.
The mapping is performed using a fresh set of current images from the moment in which the relevant action starts being executed. 
This reduces sensitivity to small scene changes occurring during VLM inference or the execution of preceding actions, increasing the robustness of the pipeline.

Once the execution is completed, another new set of images is taken and preprocessed.
These new images, along with the initial images that were used to generate the plan, are then passed to the VLM in verification mode.
The preprocessing is especially important here, as the VLM must recognize any differences between the two sets of images to ascertain whether they match the plan's expected goal.
However, in dense environments, the sheer amount of information provided to the VLM can be overwhelming.
This can lead the VLM to disregard key information by attending to the wrong masks.
Instead, we introduce a \textit{dynamic preprocessing selection} cycle.
Initially, the VLM is only provided with the unpreprocessed images, as well as the list of object labels.
Using these labels, the VLM can iteratively request additional preprocessed images until it has enough information to generate an answer.
This limits the number of simultaneous visual annotations and helps the verifier focus on task-relevant changes in object-dense scenes.
Analogously to the \textit{Thought} step in the planning mode, the VLM first describes the images, focusing on their differences, giving the model space to reason.
Then, it decides whether or not it requires more information based on these descriptions.
Finally, once the verification process is finished, the model's \textit{Observation} is appended to the VLM's input in planning mode for the next iteration of the pipeline.

%% file: figures/pipeline/pipeline.tex
\begin{tikzpicture}[pipeline]
  % \draw[step=.1cm, gray!10, very thin] (0, 0) grid (14, 10);
  % \draw[step=.5cm, gray!20, very thin] (0, 0) grid (14, 10);
  % \draw[step=1cm, gray!50, very thin] (0, 0) grid (14, 10);

\renewcommand{\ttdefault}{cmtt}
\renewcommand{\tiny}{\fontsize{6}{7pt}\selectfont}

%% --------------------------------
%% Components:
%% --------------------------------

%% VLMs

  \node[vlm, fill=Goldenrod!80] (plan) at (6.6, 9.3) {VLM\\{\tiny (Planning Mode)}};
  \node[number] at (plan.north east) {2};
  \node[vlm, fill=SeaGreen!70] (verify) at (6.6, 1) {VLM\\{\tiny (Verification Mode)}};
  \node[number] at (verify.north east) {6};

%% Robot

  \node[
    component,
    fill=Thistle!60,
    minimum width=1.6cm,
    text width=1.4cm,
    minimum height=2.2cm
  ] (act) at (12.2, 5) {\includegraphics[width=1.4cm]{
    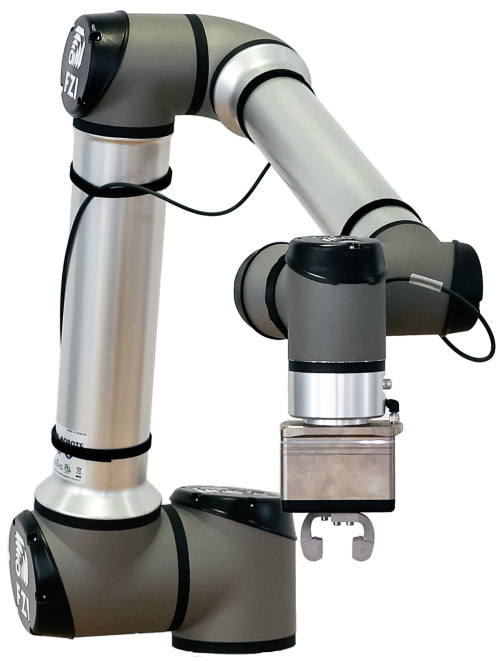
  }};
  \node[label] at (act.north west) {Robot\strut};
  \node[number] at (act.north east) {4};

%% Vision

  \node[
    component,
    fill=CornflowerBlue!60,
    minimum width=2.6cm,
    text width=2.4cm,
    minimum height=1.6cm
  ] (vision) at (6.6, 5) {Vision\\Preprocessing\\Module};

  \node[photo, anchor=center] (camera) at (6.6, 7.2) {
    \includegraphics[width=2.2cm]{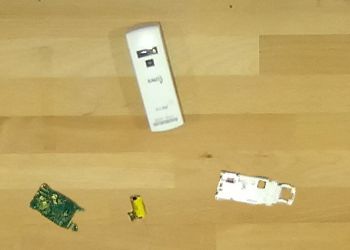}
  };
  \node[label] at (camera.north west) {Camera\strut};

%% --------------------------------
%% Data:
%% --------------------------------

%% Planning Prompt

  \node[
    group,
    minimum width=4.7cm,
    minimum height=6.2cm
  ] (planning input) at (0, 5.4) {};
  \node[label, yshift=1pt] at (planning input.north west) {Planning Prompt\strut};
  \node[number] at (planning input.north east) {1};

  \node[
    data, 
    minimum height=1.3cm,
    minimum width=4.3cm,
    text width=4.1cm
  ] (instructions) at (.2, 8.2) {\texttt{Your task is to disassemble a heat cost allocator. 
  The main goal is the removal of the battery, by detaching it from the heat cost allocator's...}};
  \node[label] at (instructions.north west) {Task Instructions\strut};

  \node[
    data, 
    vision output,
    minimum height=.6cm,
    minimum width=2.2cm
  ] (planning labels) at (2.3, 6.5) {\texttt{Case, Battery, ...}};
  \node[label] at (planning labels.north west) {Object Labels\strut};

  \node[photo] (image) at (2.3, 4) {
    \includegraphics[width=2.2cm]{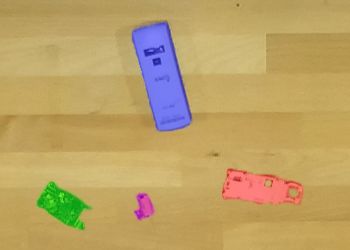}
  };
  \node[label] at (image.north west) {Image\strut};

  \node[
    data, 
    minimum height=2.1cm,
    minimum width=1.9cm,
    text width=1.7cm
  ] (actions) at (.2, 6.5) {\texttt{open\_gripper()\\
    close\_gripper()\\
    move\_to()\\
    move\_next\_to()\\
    move\_towards()\\
    mill()\\
    ...
  }};
  \node[label] at (actions.north west) {Actions\strut};

  \node[
    data, 
    checking output,
    minimum height=1.55cm,
    minimum width=1.9cm,
    text width=1.7cm
  ] (memory) at (.2, 4) {\textit{Observation 1\\
    Observation 2\\
    Observation 3\\
    Observation 4\\
    ...
  }};
  \node[label] at (memory.north west) {Memory\strut};

%% Sequence

  \node[
    data, 
    planning output,
    minimum height=1.6cm,
    minimum width=1.9cm,
    text width=1.7cm
  ] (sequence) at (11.1, 9.2) {\texttt{open\_gripper()\\
    move\_to(case)\\
    close\_gripper()\\
    move\_to(home)\\
    ...
  }};
  \node[label] at (sequence.north west) {Plan\strut};
  \node[number] at (sequence.north east) {3};

%% Checking Prompt

  \node[
    group,
    minimum width=5cm,
    minimum height=3cm
  ] (checking input) at (9, 1.5) {};
  \node[label, yshift=1pt] at (checking input.north west) {Verification Prompt\strut};
  \node[number] at (checking input.north east) {5};

  \node[
    data, 
    vision output,
    minimum height=.6cm,
    minimum width=2.2cm
  ] (checking labels) at (9.2, 2.7) {\texttt{Case, Battery, ...}};
  \node[label] at (checking labels.north west) {Object Labels\strut};

  \node[
    data, 
    planning output,
    minimum height=.6cm,
    minimum width=2.2cm
  ] (goal) at (11.6, 2.7) {\texttt{"Picked up case?"}};
  \node[label] at (goal.north west) {Goal\strut};

  \node[photo] (before) at (9.2, 1.7) {
    \includegraphics[width=2.2cm]{figures/pipeline/before_compressed}
  };
  \node[label] at (before.north west) {Before\strut};

  \node[photo] (after) at (11.6, 1.7) {
    \includegraphics[width=2.2cm]{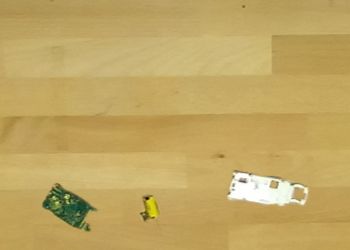}
  };
  \node[label] at (after.north west) {After\strut};

%% Observation

  \node[
    data, 
    checking output,
    minimum height=.6cm,
    minimum width=3.5cm,
    text width=3.3cm
  ] (observation) at (.6, 1.3) {\texttt{"Case picked up successfully."}};
  \node[label] at (observation.north west) {Observation\strut};
  \node[number] at (observation.north east) {7};

%% --------------------------------
%% Arrows:
%% --------------------------------

  \draw[arrow] (camera.south) -- (vision.north);

%% Planning outputs

  \draw[arrow] ($(plan.south east) + (0, .3)$) -- ($(sequence.north west) + (0, -.3)$);
  \draw[arrow] ($(plan.north east) + (0, -.3)$) -| ($(goal.north east) + (-.3, 0)$);
  \draw[arrow] ($(sequence.south) + (.15, 0)$) -- (act.north);

%% Acting outputs

  \draw[arrow] (act.south) -- ($(checking input.north) + (.7, 0)$);
  \draw[double arrow] (act.west) -- (vision.east)
  node[arrow label] {%
    \texttt{generate\_pose(object)}
  };
  \draw[arrow] ($(checking input.west) + (0, -.5)$) -- (verify.east);

%% Checking outputs

  \draw[arrow] (verify.west) -- (observation.east);
  \draw[arrow] (observation.north) -- (planning input.south);
  \draw[arrow] (planning input.north) |- (plan.west);

%% Vision outputs

  \draw[
    arrow,
    font=\ttfamily,
    transform shape
  ][postaction={
    decorate,
    decoration={
      text effects along path,
      text={~mask\_objects},
      text align={center},
      reverse path,
      text effects/.cd,
        character 6/.style={yshift=-.48ex},
    }
  }] ($(vision.south west) + (0, .4)$) .. controls ($(vision.south west) + (-1.4, .4)$) .. ($(image.north) + (-.2, 0)$);

  \draw[
    arrow,
    font=\ttfamily,
    transform shape
  ][postaction={
    decorate,
    decoration={
      text effects along path,
      text={~find\_objects},
      text align={center},
      reverse path,
      text effects/.cd,
        character 6/.style={yshift=-.48ex},
    }
  }] ($(vision.north west) + (0, -.4)$) .. controls ($(vision.north west) + (-1.4, -.4)$) .. ($(planning labels.south) + (-.2, 0)$);

%% DPS

  \draw[
    arrow,
    font=\ttfamily,
    transform shape
  ][postaction={
    decorate,
    decoration={
      text effects along path,
      text={mask\_objects~},
      text align={center},
      text effects/.cd,
        character 5/.style={yshift=-.48ex},
    },
  }] ($(verify.north) + (-.7, .2)$) .. controls (5.3, 2.5) and (5.3, 3.5) .. ($(vision.south) + (-.7, 0)$);
  
  \draw[arrow] ($(vision.south) + (.7, -.2)$) .. controls (7.9, 3.5) and (7.9, 2.5) .. ($(verify.north) + (.7, 0)$);

  \node[text width=2cm, text=darkgray, align=center] (dps) at ($(vision.center) + (0, -2)$) {Dynamic\\Preprocessing\\Selection};

  \node[photo, anchor=center] (camera) at ($(dps.center) + (1.5, 0)$) {
    \includegraphics[width=1.2cm]{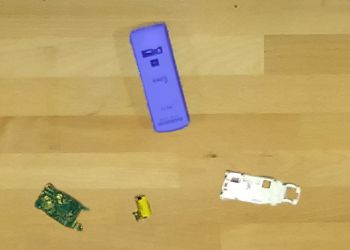}
  };

  % \fontsize{130}{120}
  % \node[text=red, opacity=0.2] at (6.5, 5) {\textbf{W I P}};
\end{tikzpicture}

%% file: figures/toolbox.tex
\begin{tabular}{l l}
    \hline
    \textbf{Action}         & \textbf{Parameters}           \\
    \hline
    \texttt{open\_gripper}  & N/A                           \\
    \texttt{close\_gripper} & N/A                           \\
    \texttt{move\_to}       & location label                \\
    \texttt{move\_next\_to} & location label \& direction   \\
    \texttt{move\_towards}  & direction                     \\
    %\hline
    \multicolumn{2}{l}{Additional task-specific actions\dots} \\
    \hline
\end{tabular}

%% file: figures/tasks.tex
\begin{tabular}{l l l l c}
    \hline
                & \textbf{Task}                     & \textbf{Objects}      & \textbf{Domain}   & \textbf{Evaluation}   \\
    \hline
    \textbf{A}  & PC assembly                       & PC components         & Workshop          & Simulation            \\
    \textbf{B}  & Crop care \& harvest              & Plant species         & Agriculture       & Simulation            \\
    \textbf{C}  & Order fulfillment                 & Facial recognition    & Service           & Simulation            \\
    \textbf{D}  & Cockpit interaction               & Button/dial status    & Aerospace         & Simulation            \\
    \textbf{E}  & Heat cost allocator disassembly   & Allocator parts       & E-waste recycling & Hardware              \\
    \hline
\end{tabular}

%% file: sections/experiments.tex
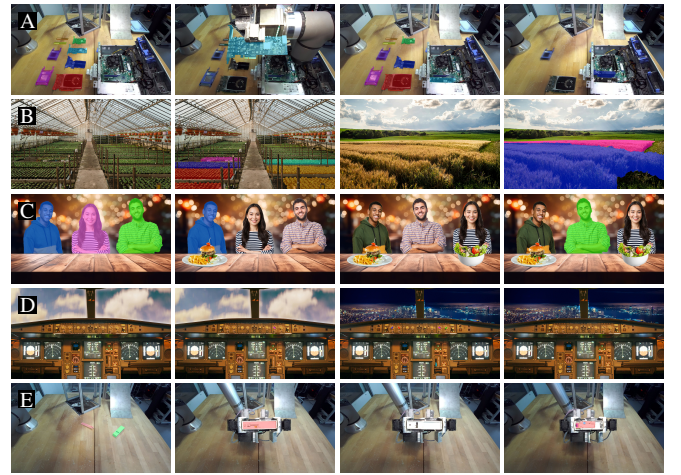
\begin{figure}[hb]
    \input{figures/collage}
    \caption{Example excerpts of input image sequences for the VLM, as used in the evaluation.
    The sequences are labeled A through E, corresponding to the tasks in Table~\ref{tab:tasks}.}
    \label{fig:collage}
\end{figure}

\section{Experimental Setup}
\label{sec:experiments}

The evaluation addresses three questions: whether explicit domain-specific visual grounding improves VLM-based planning and verification, whether visual preprocessing provides benefits beyond textual domain descriptions, and whether VLM-directed preprocessing helps verification in object-dense scenes.
To investigate these questions, we compare five configurations constituting ablations of the grounding interface of the cyclic planning pipeline.
Initially, we evaluate the performance of the pipeline without the \textit{modular vision preprocessing} (and consequently, without the \textit{dynamic preprocessing selection}).
This forms our baseline.
The second configuration augments the basic VLM inputs with natural language descriptions ($D$) of the domain-specific objects that can appear in the environment. 
Next, we introduce the \textit{modular vision preprocessing} ($V$) to the pipeline, both with and without the natural language descriptions.
Finally, we evaluate the full pipeline with the \textit{dynamic preprocessing selection} ($S$), in order to assess its effect on the scalability of the preprocessing in object-dense environments.

The evaluation consists of a controlled image-sequence study and a real-robot case study.
Both provide the VLM with the same forms of visual and textual input, allowing the planning and verification interface to be evaluated consistently.
The controlled study isolates the effect of the grounding information, whereas the hardware study demonstrates how the complete system can be instantiated using learned perception, conventional pose estimation, and an established robot execution stack.

\subsection{Evaluation Tasks}
\label{subsec:experiments_tasks}
In order to evaluate the breadth of the pipeline's applicability, we consider five different tasks from specialized domains, as summarized in Table~\ref{tab:tasks}.
The tasks were chosen to cover a wide range of scenarios, include different types of objects, and require multiple interaction steps to complete.
To help visualize the tasks, example input images are shown in Fig.~\ref{fig:collage}.

In task \textbf{A}, the robot is required to pick up a series of PC components (\eg GPU, RAM, SSD, etc.) and install them into a computer case, according to the given specification.
If a given slot is already occupied, the robot must first remove the existing component before proceeding.
A core challenge of this task is recognizing which components are installed in the computer's case in any given moment.

Task \textbf{B} requires the robot to recognize specific crops and handle them in accordance with the instructions, such as harvesting, watering or fertilizing them.
Both greenhouse plants and field crops are considered.
The main challenge of this task is the fine-grained recognition of different plant species, such as distinguishing between wheat and barley.

Task \textbf{C} is a service task, in which the robot must accept an order and serve it to the correct customer.
The customers can be differentiated using facial recognition.

In task \textbf{D}, the robot must complete a simplified flying sequence by interacting with the cockpit of a small aircraft.
The task investigates the scalability of the approach, given the large number of interactable objects.

Finally, for task \textbf{E} the robot must disassemble different variants of heat cost allocators, the main goal being the removal of the built-in battery.
Destructive means of disassembly are allowed and provided through the use of a milling spindle, which can be interacted with through a dedicated action.
Being executed on a real robot, the task explores the real-world applicability of the approach.

Each task is evaluated over 11 different sequences, with variations either in the initial setup or during execution (\eg failures in the robot's actions).

\subsection{Simulation}
\label{subsec:experiments_sim}
The evaluation of tasks \textbf{A} through \textbf{D} is performed with simulated perception modules.
For each time step, the input images are manually annotated with the labels and segmentation masks that would be produced by a domain-specific perception system.
Additional alternate inputs are also created, including failed actions and unexpected scene changes.
This isolates the effect of the grounding interface on VLM planning and verification without requiring a complete perception and manipulation implementation for every domain.
It does not, however, model recognition or segmentation errors within the perception modules.

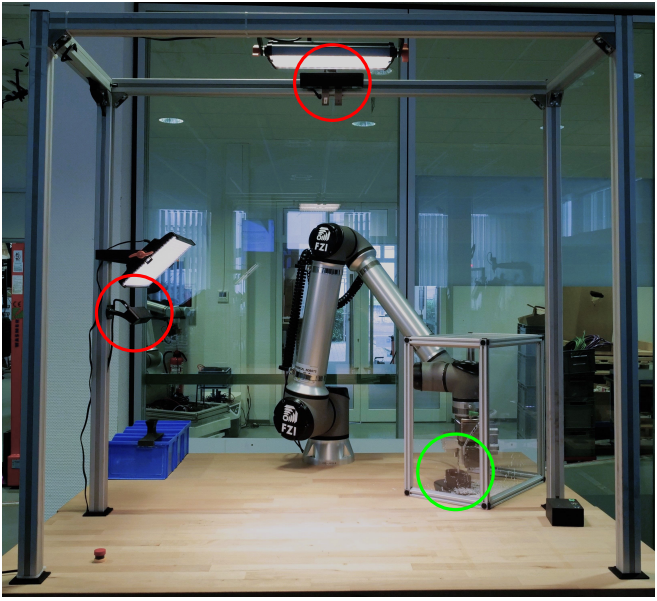
\begin{figure}[ht]
    \centering
    \vspace{1ex}
    \input{figures/setup}
    \caption{The hardware setup used for the real-world evaluation of the pipeline.
    Note the milling spindle on the right side of the table (circled in green), as well as the cameras mounted on the frame on the top and on the left (circled in red), providing both a top-down and a lateral perspective of the workspace.}
    \label{fig:setup}
\end{figure}

\subsection{Real-World Robot}
\label{subsec:experiments_real}
As seen in Fig.~\ref{fig:setup}, the evaluation environment for task~\textbf{E} consists of a table featuring a mounted Universal Robots UR10e with a Weiss Robotics CRG 200-085 gripper with custom fingers for object manipulation.
Vision is provided by two StereoLabs ZE~2i stereo cameras with inherent stereo depth estimation capabilities.
Movement trajectory planning and execution is done using the ROS 2 middleware~\cite{macenski2022robot} and the MoveIt 2 motion planning framework~\cite{coleman2014reducing}.
The BT generation and execution is handled by the \texttt{ros2\_ros\_bt\_py} framework~\cite{heppnerDistributedBehaviorTreesHeterogeneous2023}.
This implementation reuses the existing execution stack, confining application-specific adaptation primarily to the perception models and task-specific BT nodes.

The object and part recognition (as provided by the \texttt{find\_objects} function) is performed using a YOLOv9 model~\cite{wang2024yolov9learningwantlearn}, fine-tuned on a small dataset of 181 labeled instances.
For \texttt{mask\_objects}, component segmentation was not learned but is calculated using a Segment Anything 2~\cite{ravi2024sam2segmentimages} instance utilizing the bounding boxes generated by the YOLOv9 network as region proposals.
Finally, localization of the components for \texttt{generate\_pose} is executed by projecting the detected object features into 3D space using a calibrated camera model.

In order to operate the mill, the VLM can select a part (identified by the vision module) to attempt to remove using a masked close-up image of the heat cost allocator.
Based on this selection, the vision module generates a custom milling trajectory, which can be executed by the robot arm itself, which simply moves along it above the spindle while grasping the heat cost allocator.

\subsection{Choice of VLM}
\label{subsec:experiments_vlm}
The VLM used for the bulk of the evaluation is the open-source model LLaVA-OneVision~\cite{li2024llava}, consisting of a SigLIP vision encoder~\cite{zhai2023sigmoid} and Qwen2~\cite{yang2024qwen2technicalreport} text inference backbone.
However, to examine whether the grounding interface transfers between reasoning models, we repeat a subset of the evaluation using the open-source model Gemma~3~\cite{gemmateam2025gemma3technicalreport}.
With these additional experiments we investigate the transferability of the results across different VLMs, and the extent to which the choice of VLM affects the performance of the pipeline.

\subsection{Evaluation Metrics}
\label{subsec:experiments_metrics}

We manually assess the pipeline at three levels corresponding to its planning and verification stages.
Planning Accuracy denotes the proportion of sequences for which the generated plans are judged correct, independently of subsequent verification outputs.
Verification Accuracy denotes the proportion of verification decisions matching the known execution outcome and is evaluated for correctly generated plans.
Overall Accuracy denotes the proportion of sequences for which both planning and all associated verification decisions are correct.

%% file: figures/collage.tex
%\tikzsetfigurename{collage}
\begin{tikzpicture}[collage]
    \tikzmath{
        \imagewidth = .244\linewidth;
        \imagespacing = .008\linewidth;
    }
    \node[image] (A1) at (0, 0) {
        \includegraphics[width=\imagewidth pt]{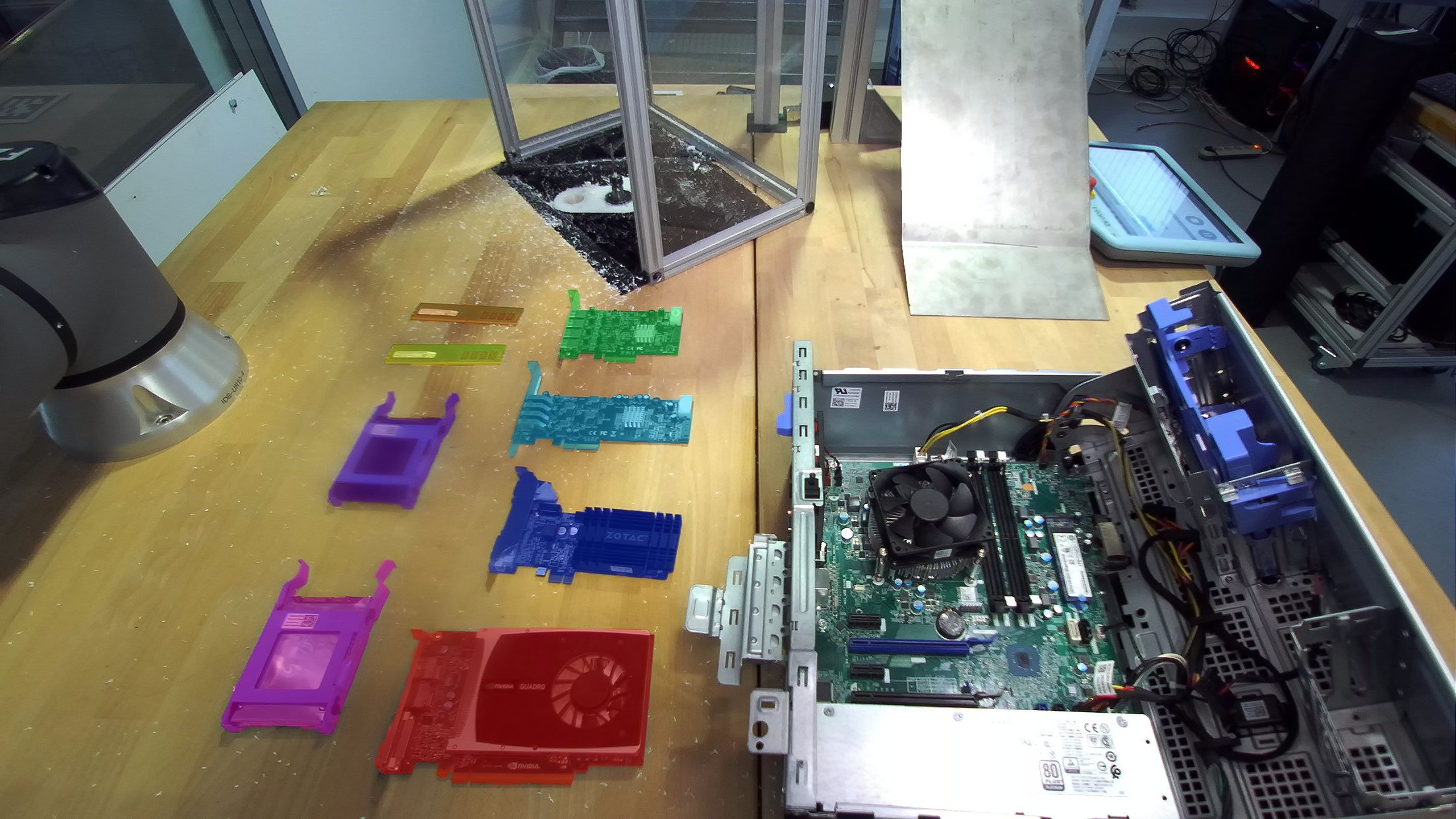}
    };
    \node[image label] at (A1.north west) {A};
    \node[image] (A2) at ($(A1.north east) + (\imagespacing pt, 0)$) {
        \includegraphics[width=\imagewidth pt]{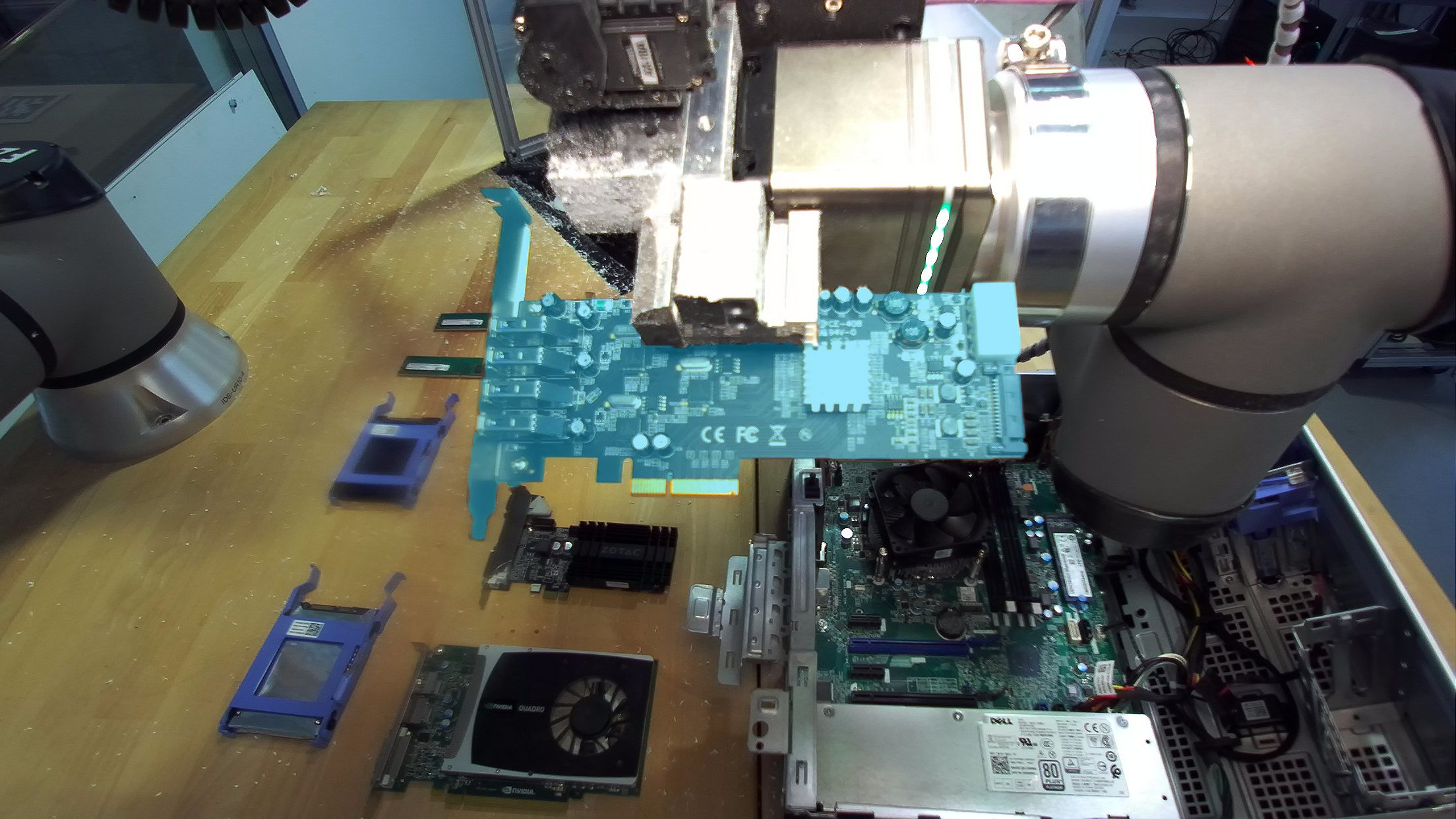}
    };
    \node[image] (A3) at ($(A2.north east) + (\imagespacing pt, 0)$) {
        \includegraphics[width=\imagewidth pt]{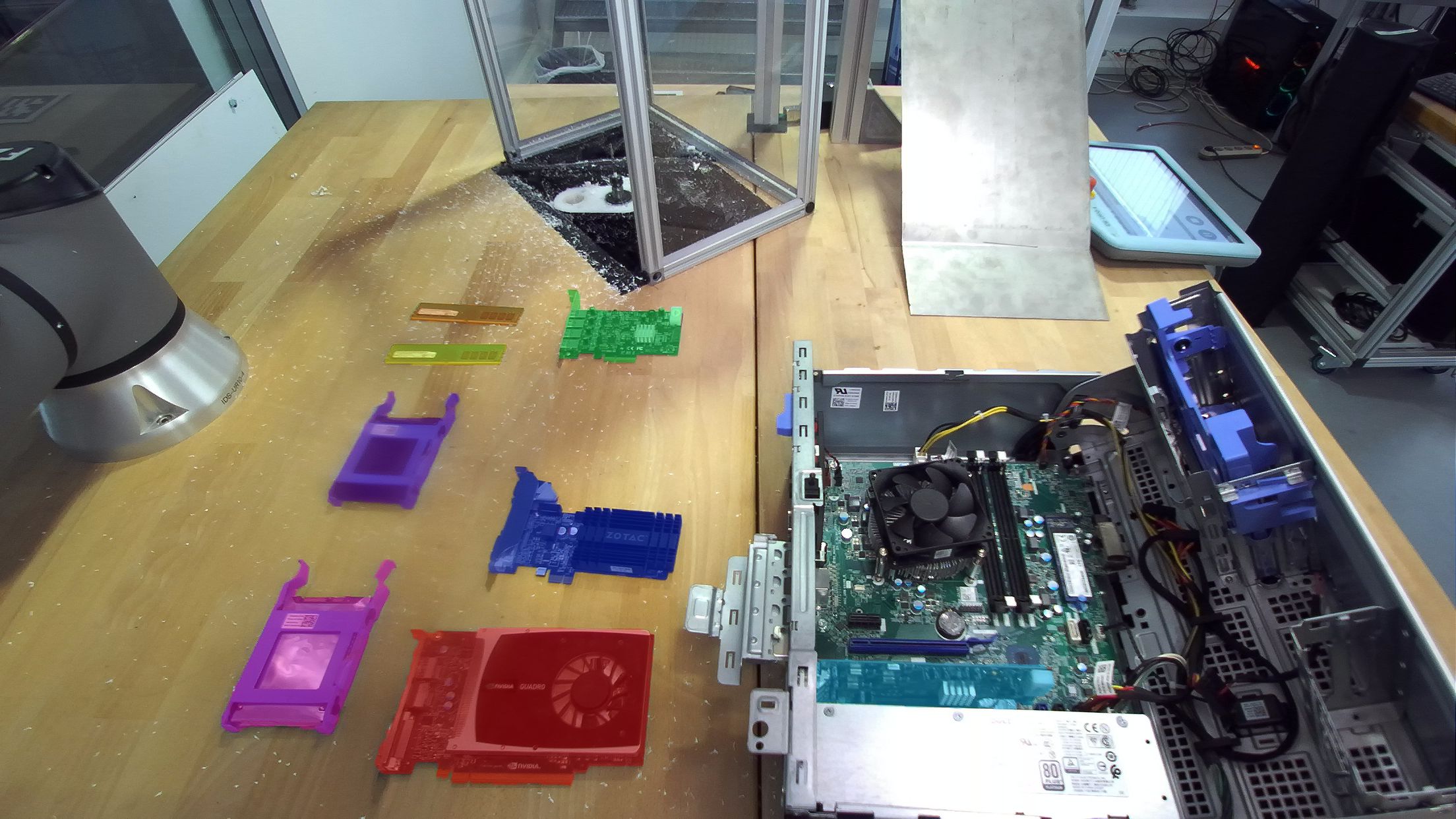}
    };
    \node[image] (A4) at ($(A3.north east) + (\imagespacing pt, 0)$) {
        \includegraphics[width=\imagewidth pt]{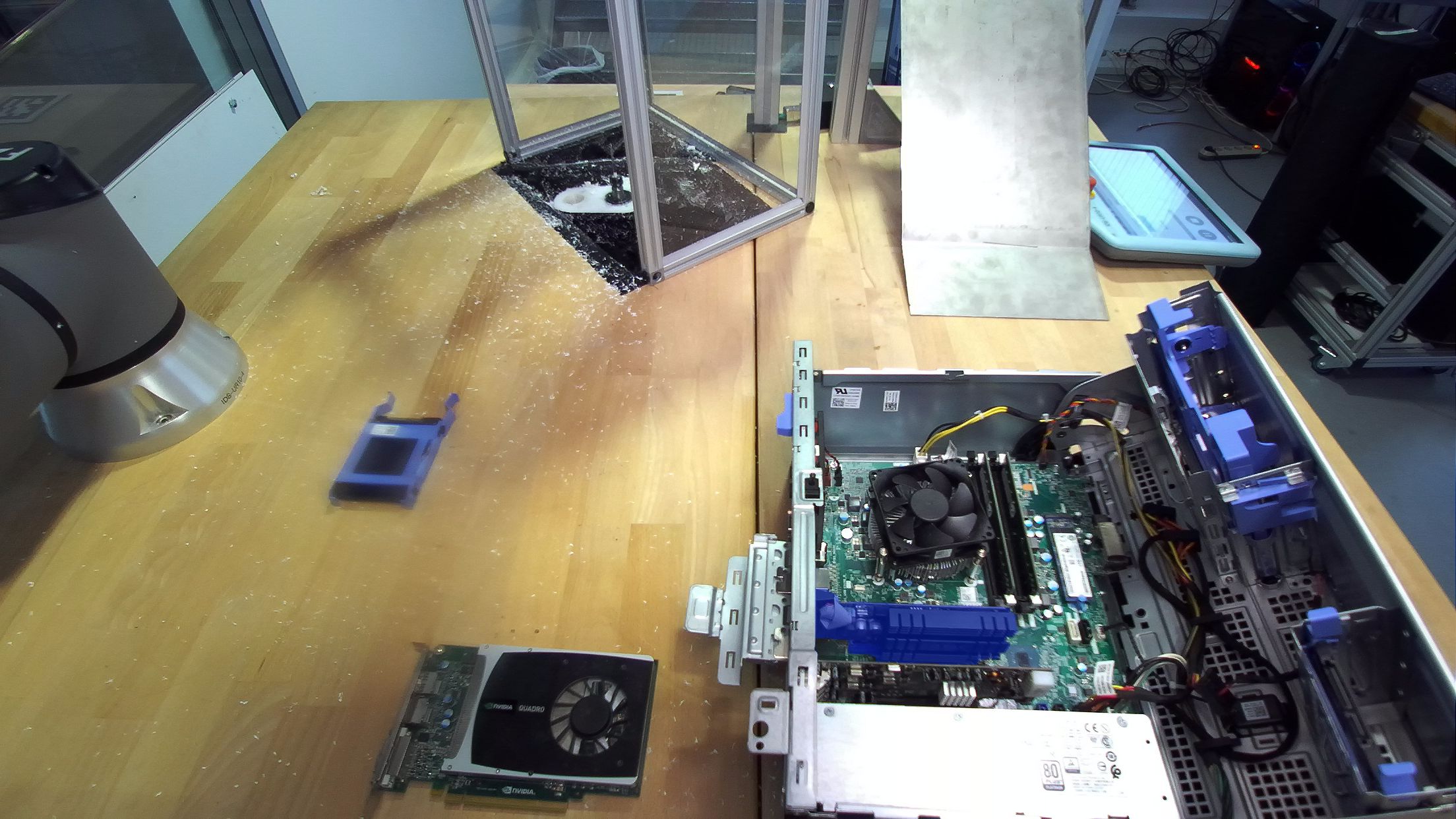}
    };

    \node[image] (B1) at ($(A1.south west) + (0, -\imagespacing pt)$) {
        \includegraphics[width=\imagewidth pt]{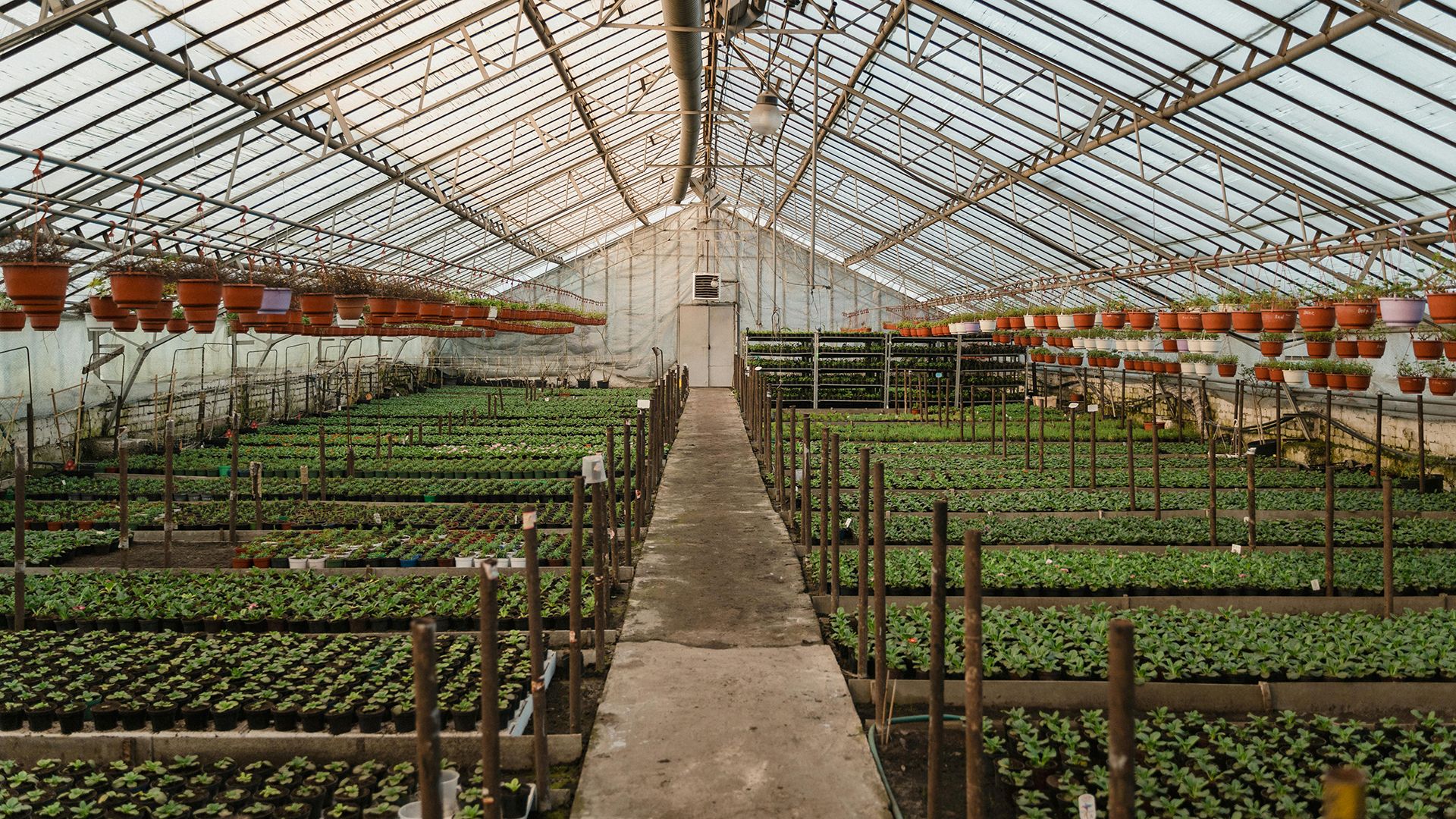}
    };
    \node[image label] at (B1.north west) {B};
    \node[image] (B2) at ($(B1.north east) + (\imagespacing pt, 0)$) {
        \includegraphics[width=\imagewidth pt]{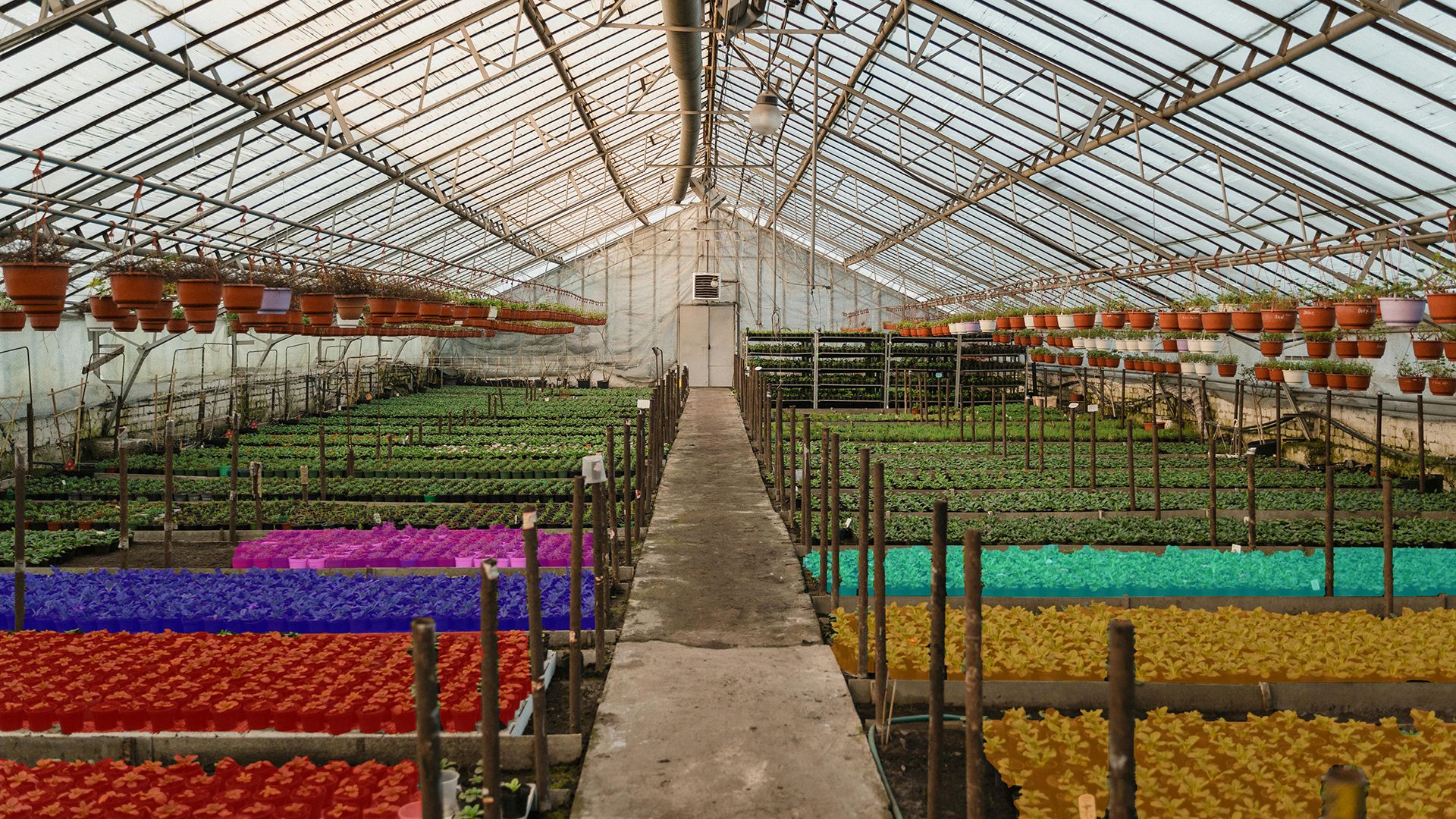}
    };
    \node[image] (B3) at ($(B2.north east) + (\imagespacing pt, 0)$) {
        \includegraphics[width=\imagewidth pt]{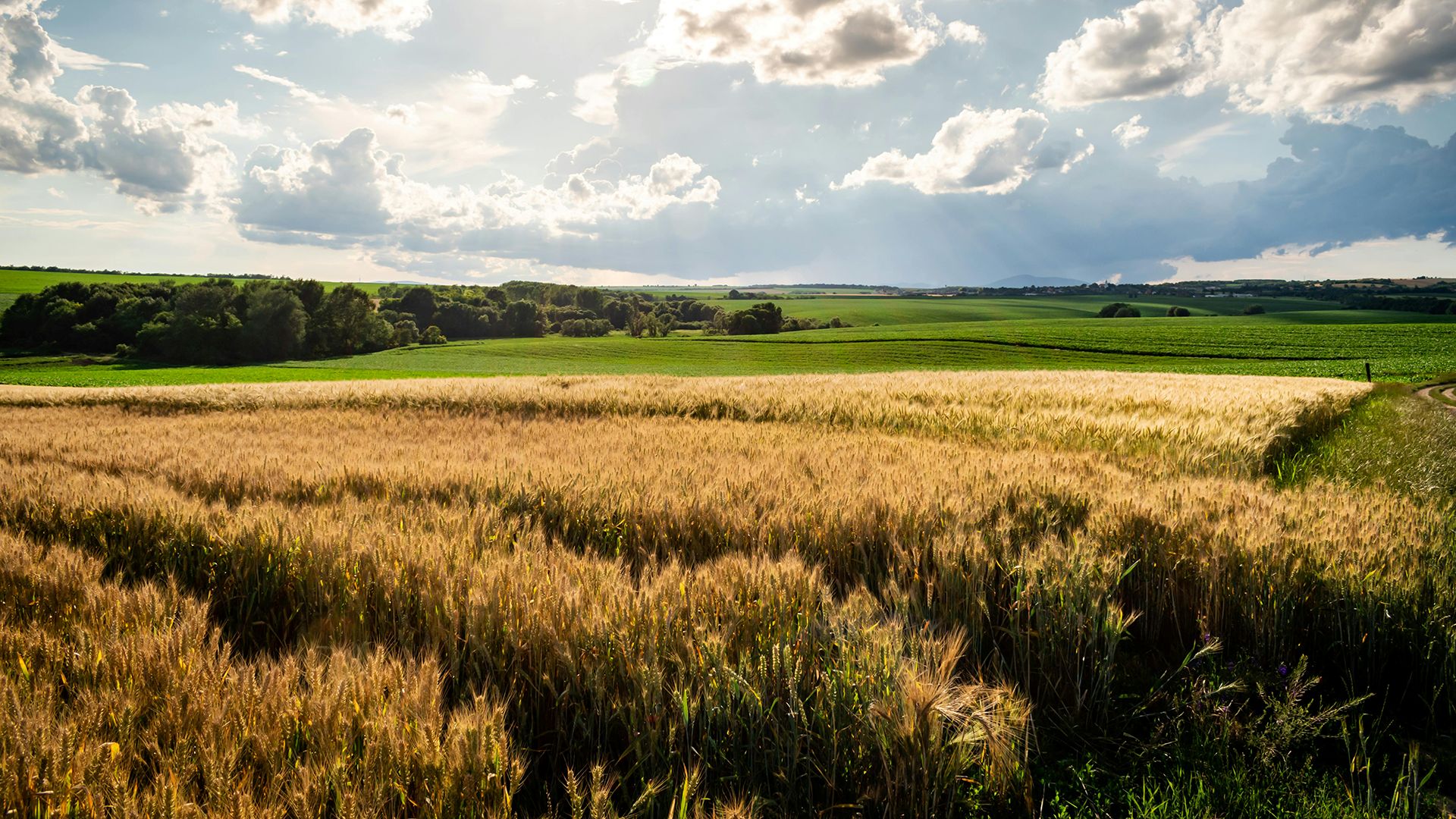}
    };
    \node[image] (B4) at ($(B3.north east) + (\imagespacing pt, 0)$) {
        \includegraphics[width=\imagewidth pt]{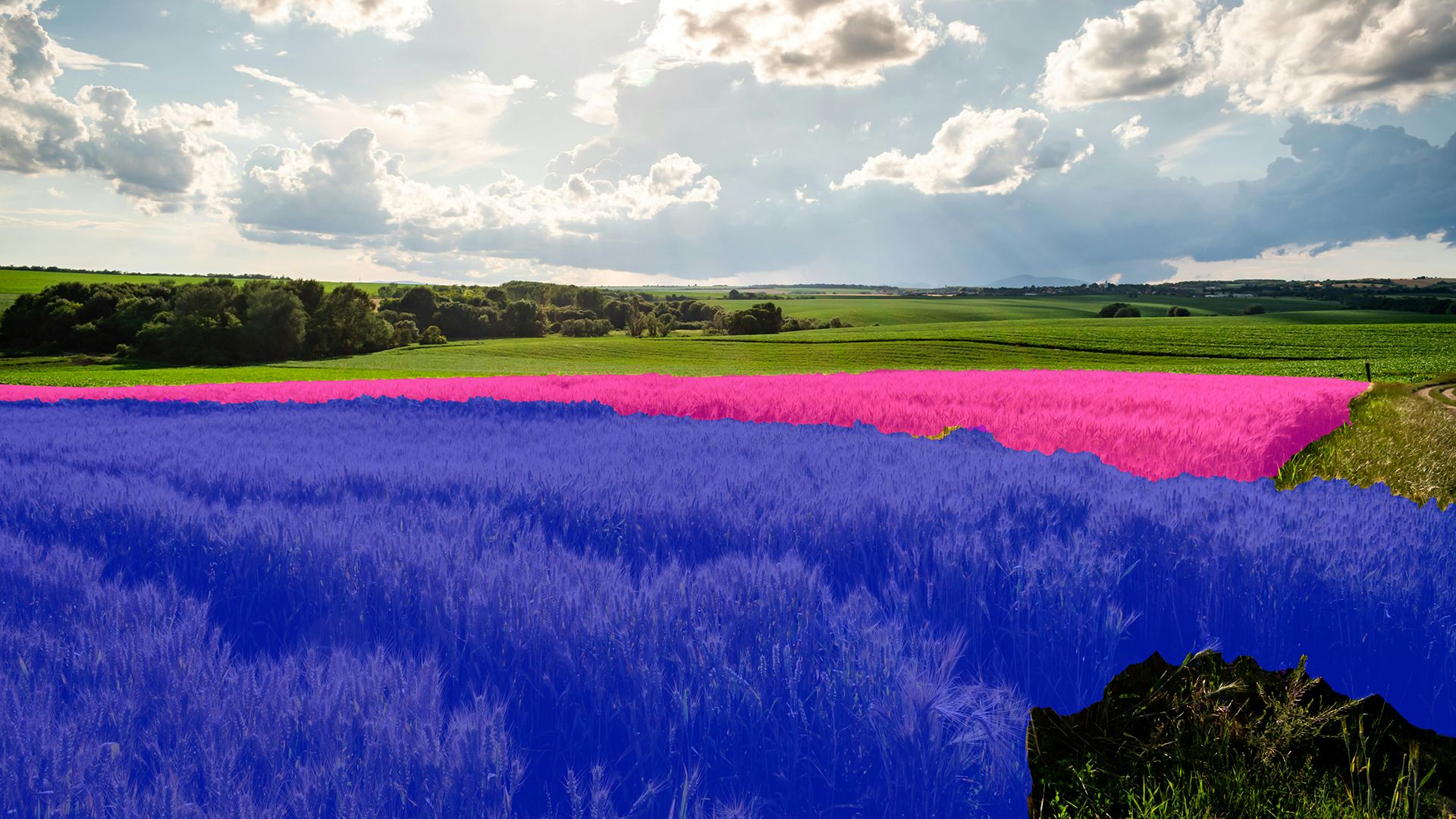}
    };

    \node[image] (C1) at ($(B1.south west) + (0, -\imagespacing pt)$) {
        \includegraphics[width=\imagewidth pt]{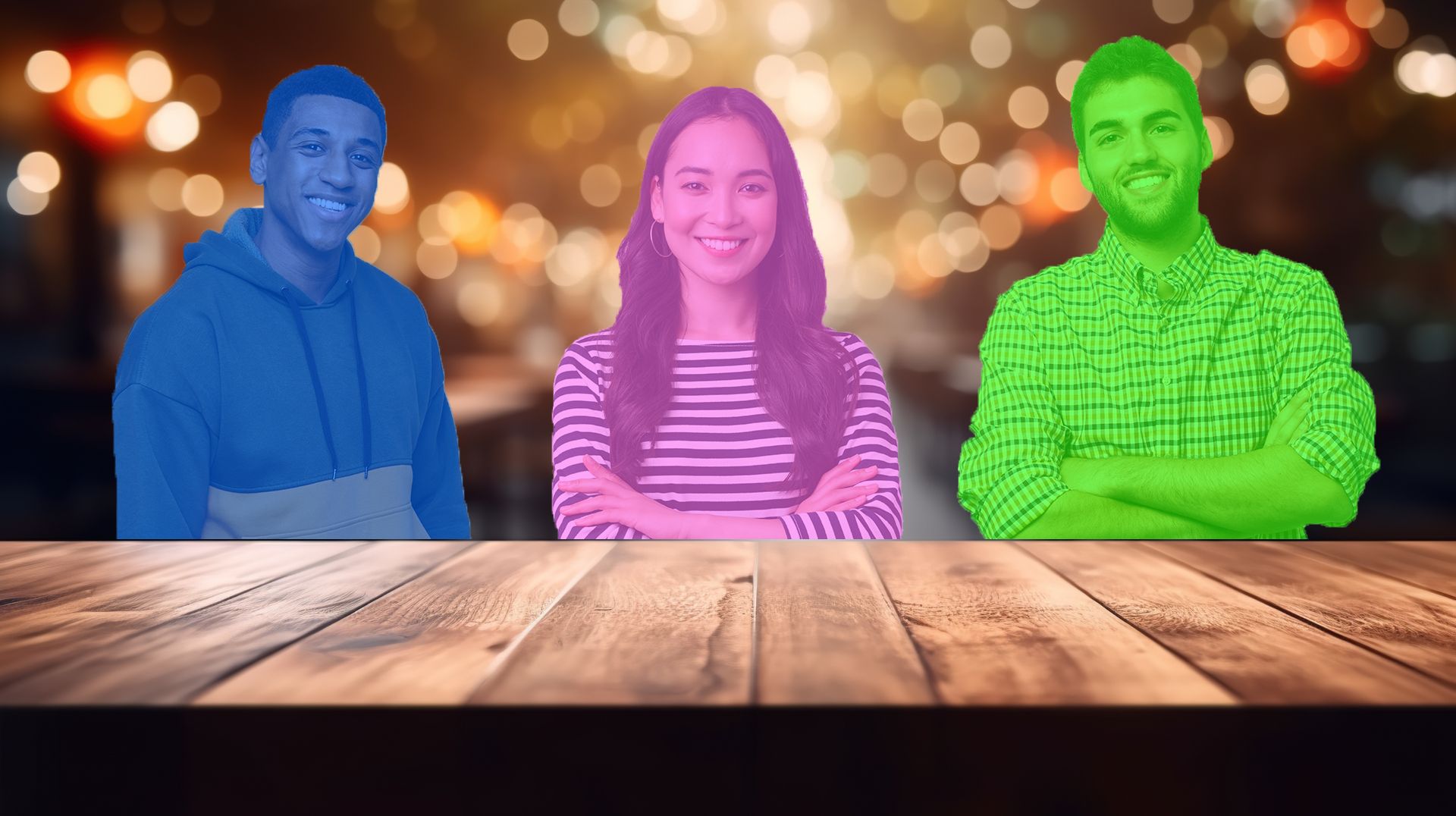}
    };
    \node[image label] at (C1.north west) {C};
    \node[image] (C2) at ($(C1.north east) + (\imagespacing pt, 0)$) {
        \includegraphics[width=\imagewidth pt]{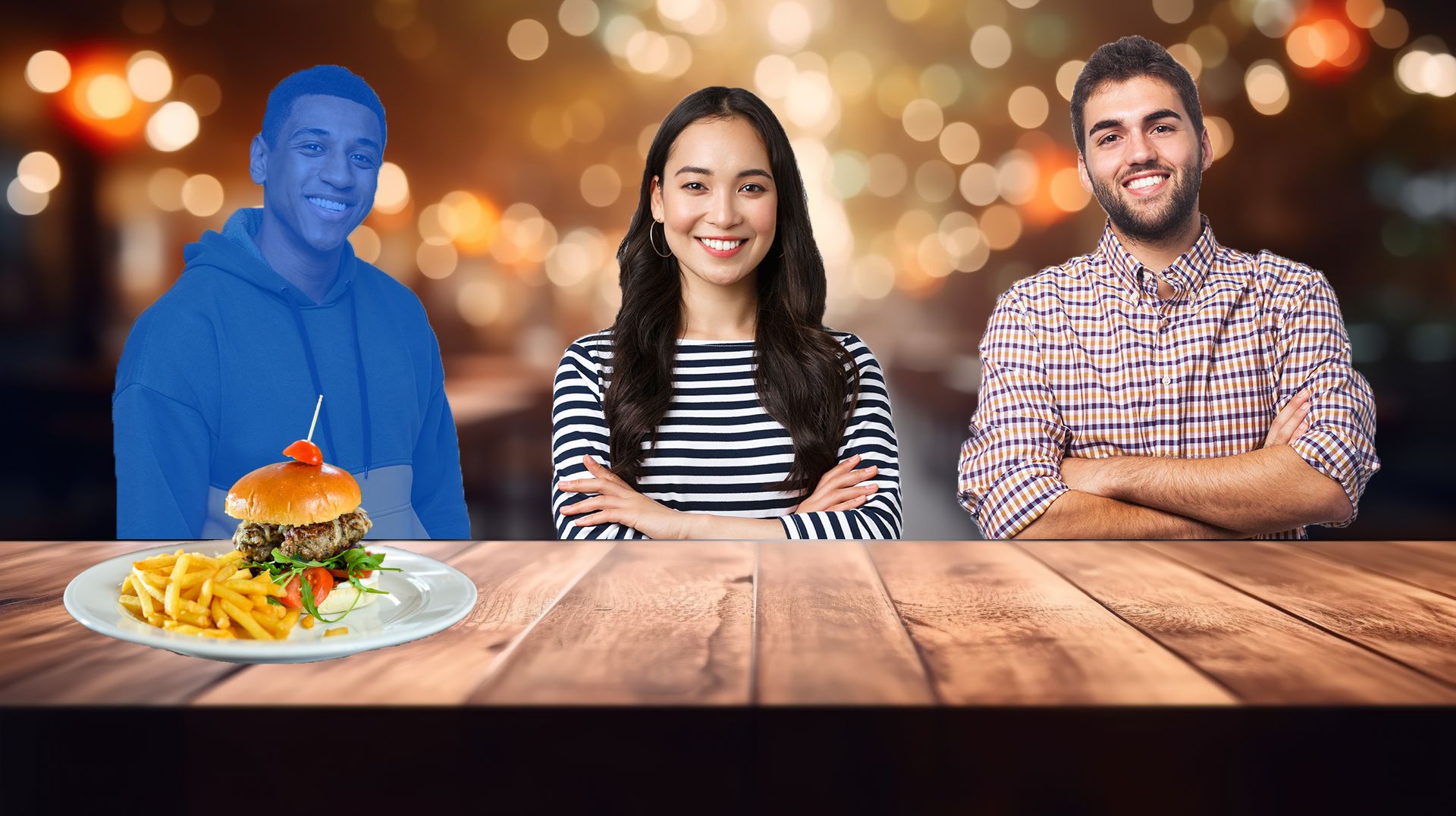}
    };
    \node[image] (C3) at ($(C2.north east) + (\imagespacing pt, 0)$) {
        \includegraphics[width=\imagewidth pt]{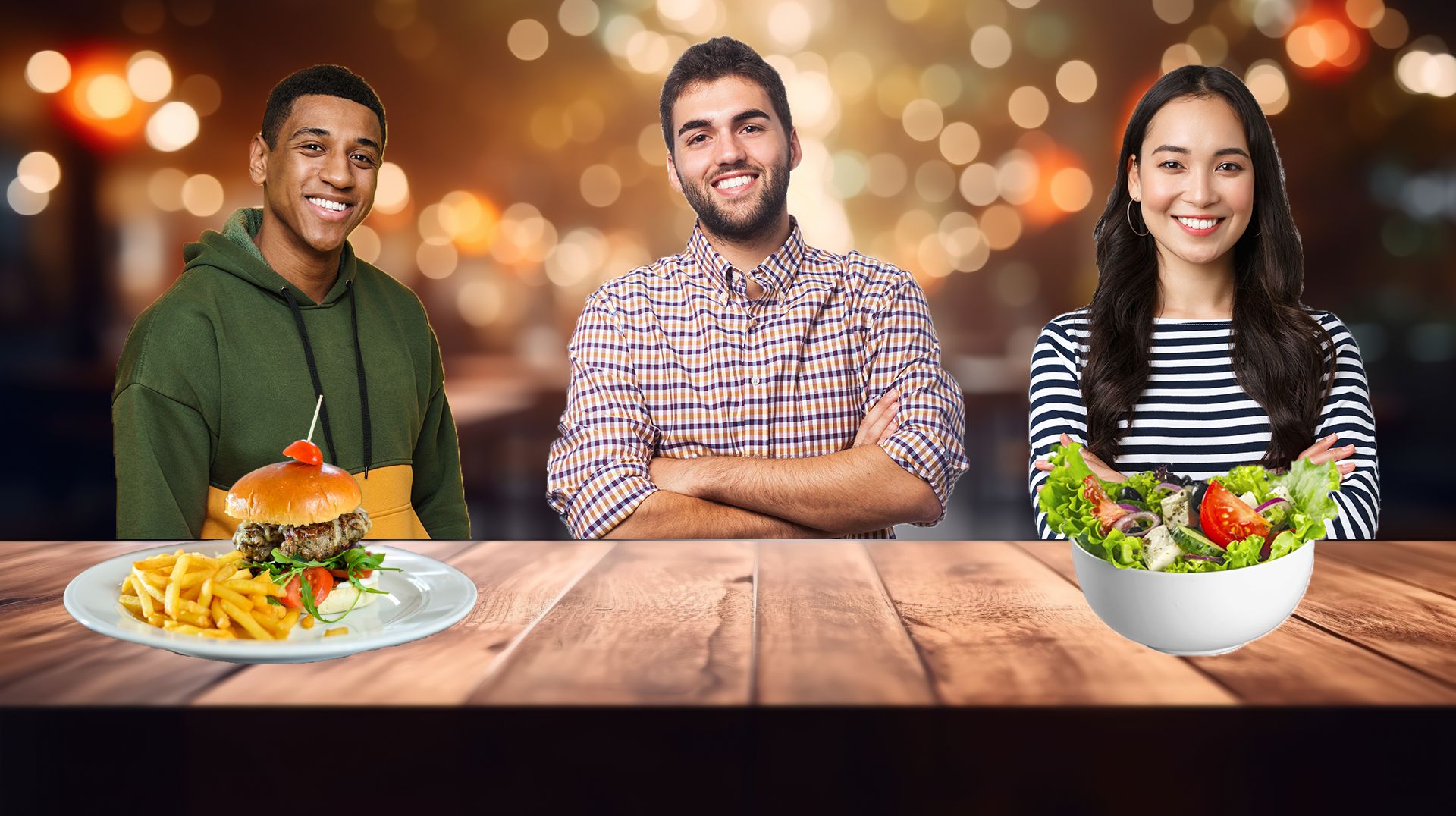}
    };
    \node[image] (C4) at ($(C3.north east) + (\imagespacing pt, 0)$) {
        \includegraphics[width=\imagewidth pt]{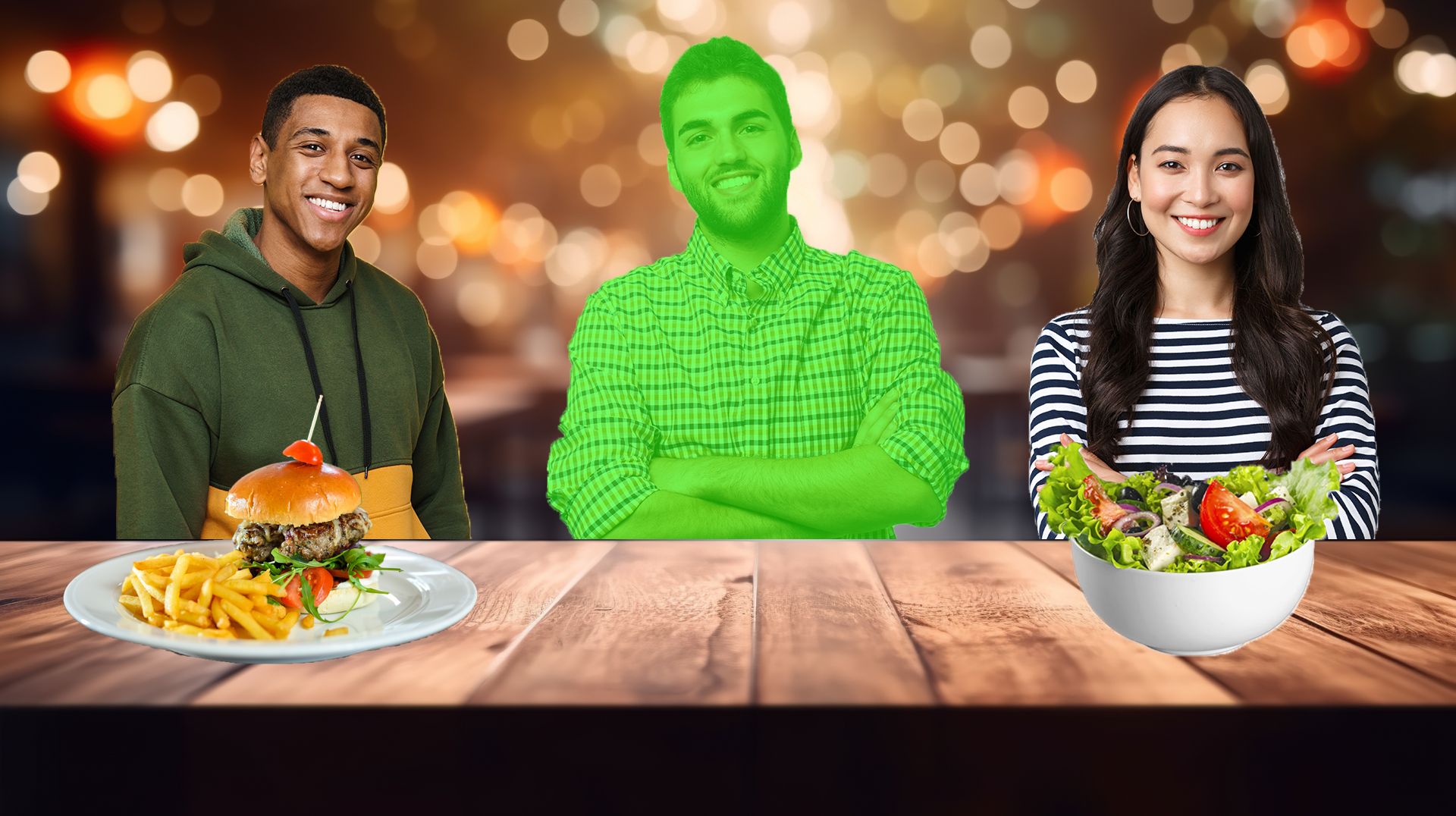}
    };

    \node[image] (D1) at ($(C1.south west) + (0, -\imagespacing pt)$) {
        \includegraphics[width=\imagewidth pt]{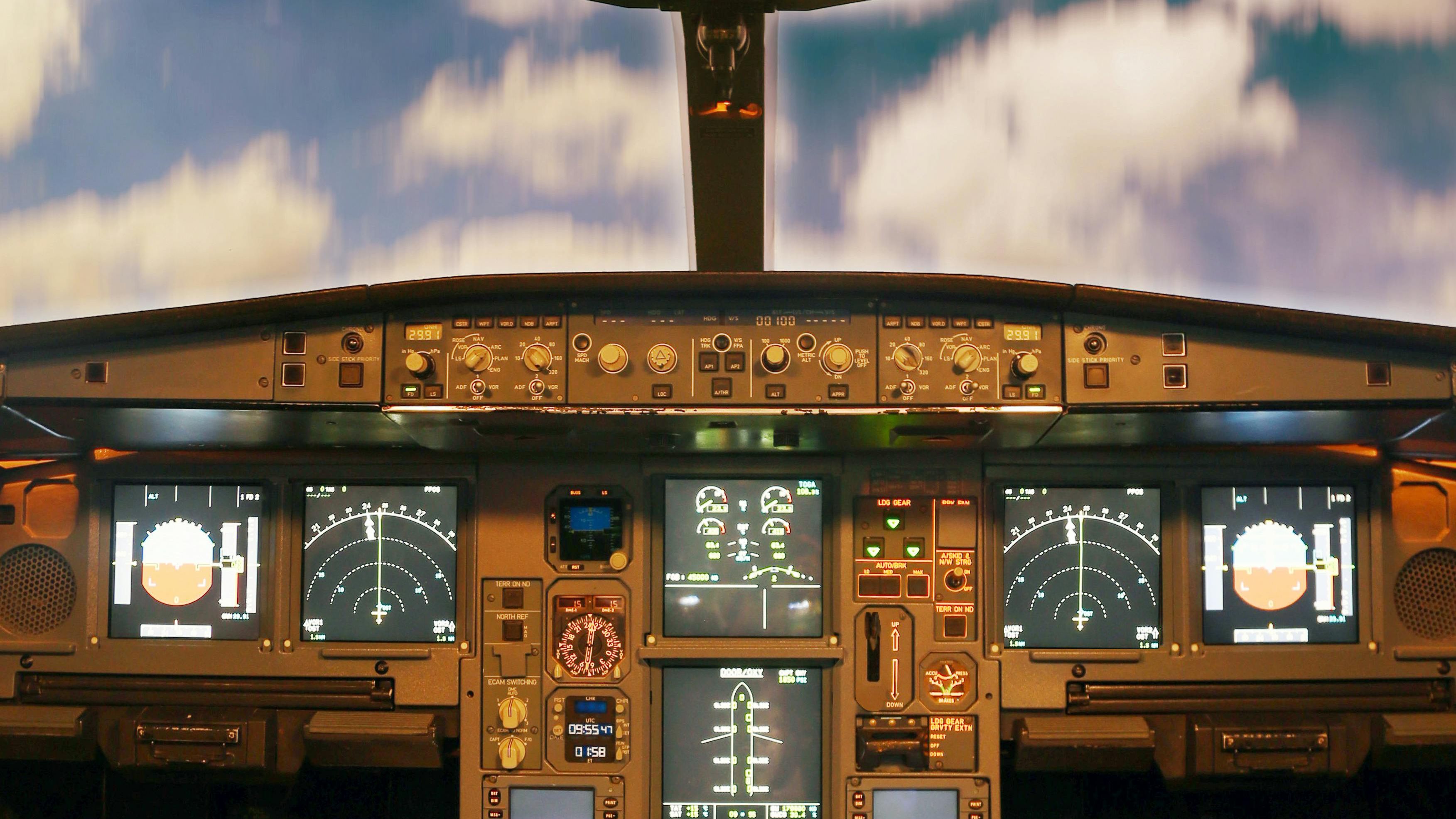}
    };
    \node[image label] at (D1.north west) {D};
    \node[image] (D2) at ($(D1.north east) + (\imagespacing pt, 0)$) {
        \includegraphics[width=\imagewidth pt]{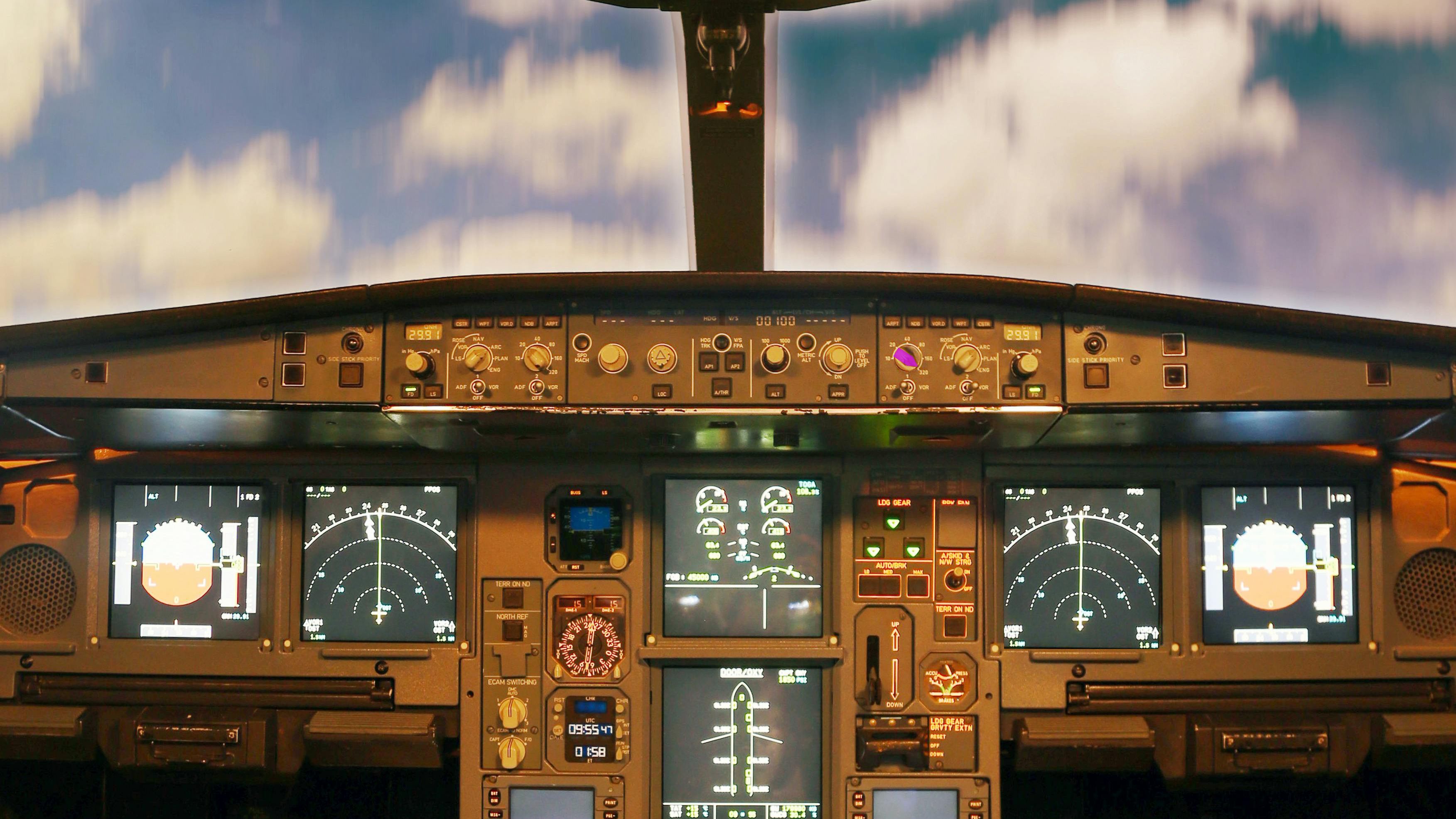}
    };
    \node[image] (D3) at ($(D2.north east) + (\imagespacing pt, 0)$) {
        \includegraphics[width=\imagewidth pt]{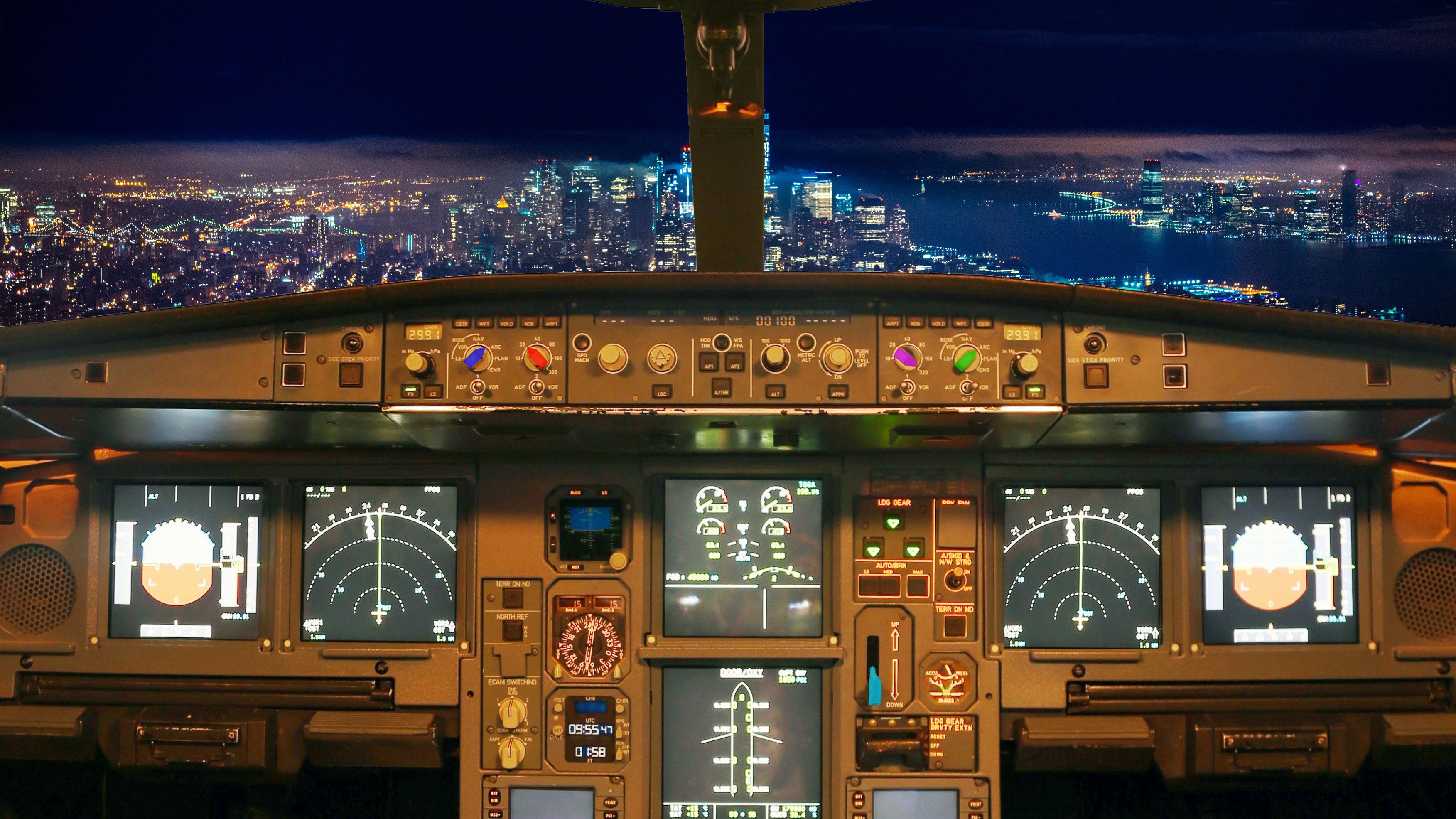}
    };
    \node[image] (D4) at ($(D3.north east) + (\imagespacing pt, 0)$) {
        \includegraphics[width=\imagewidth pt]{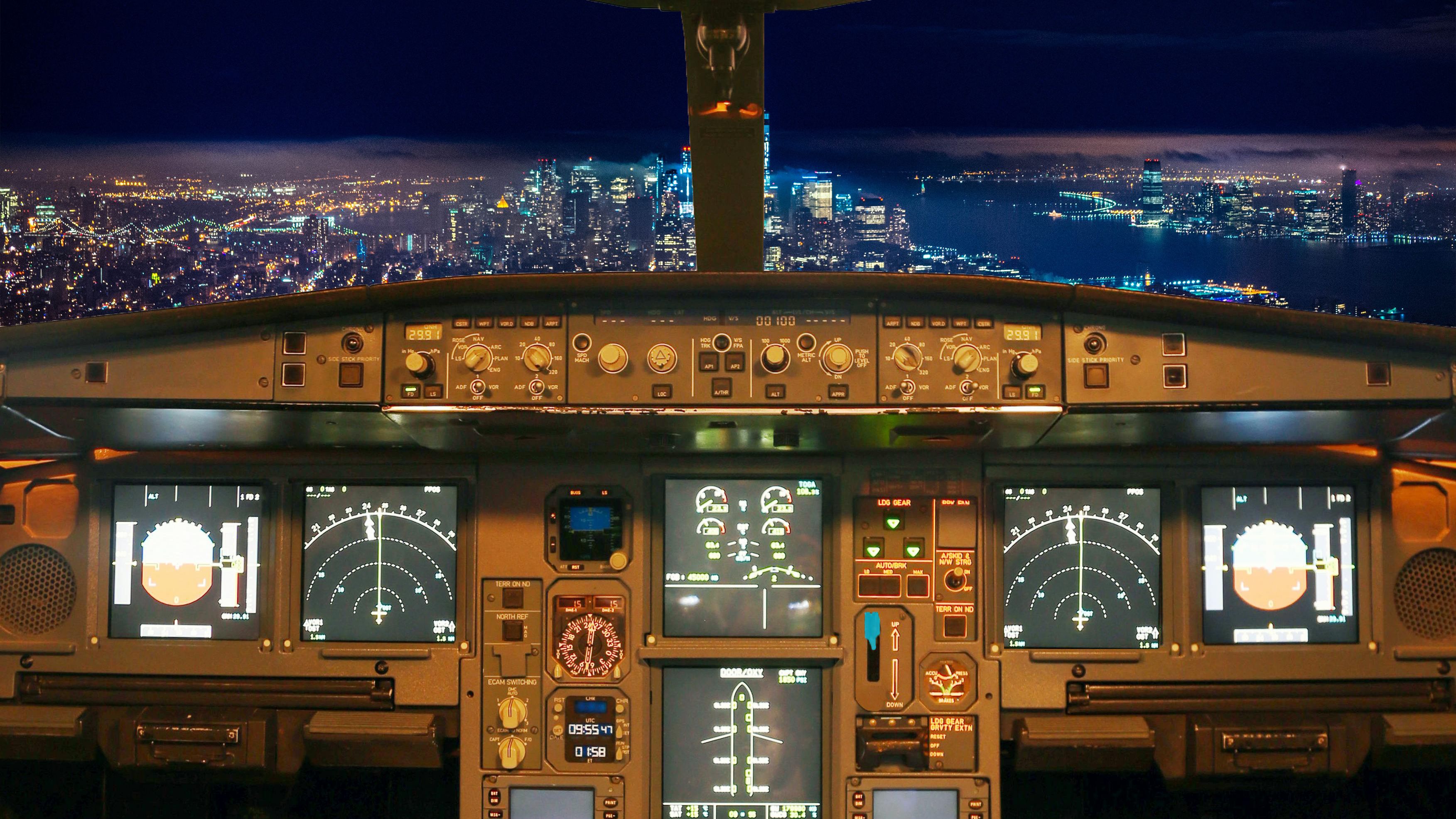}
    };

    \node[image] (E1) at ($(D1.south west) + (0, -\imagespacing pt)$) {
        \includegraphics[width=\imagewidth pt]{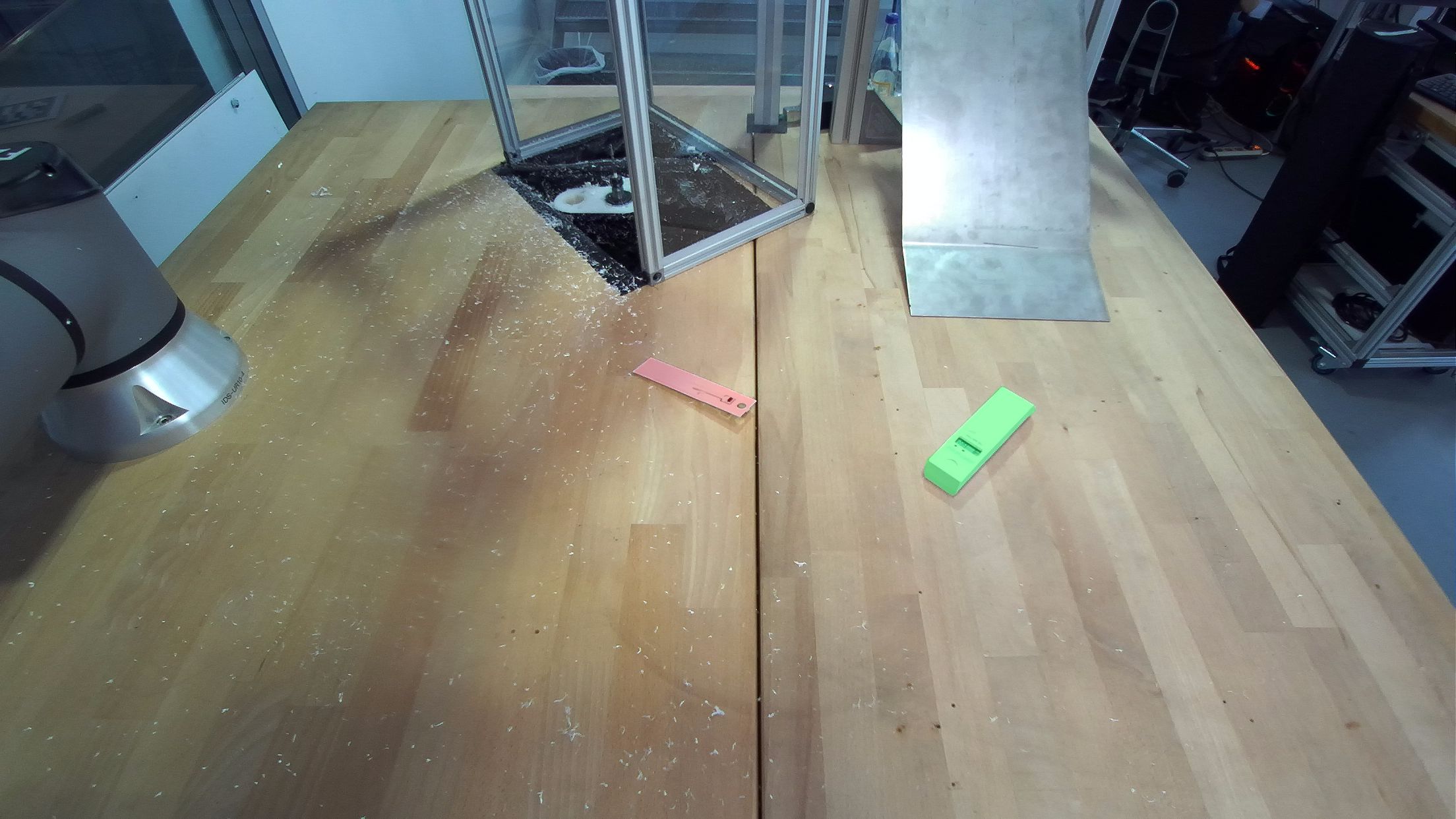}
    };
    \node[image label] at (E1.north west) {E};
    \node[image] (E2) at ($(E1.north east) + (\imagespacing pt, 0)$) {
        \includegraphics[width=\imagewidth pt]{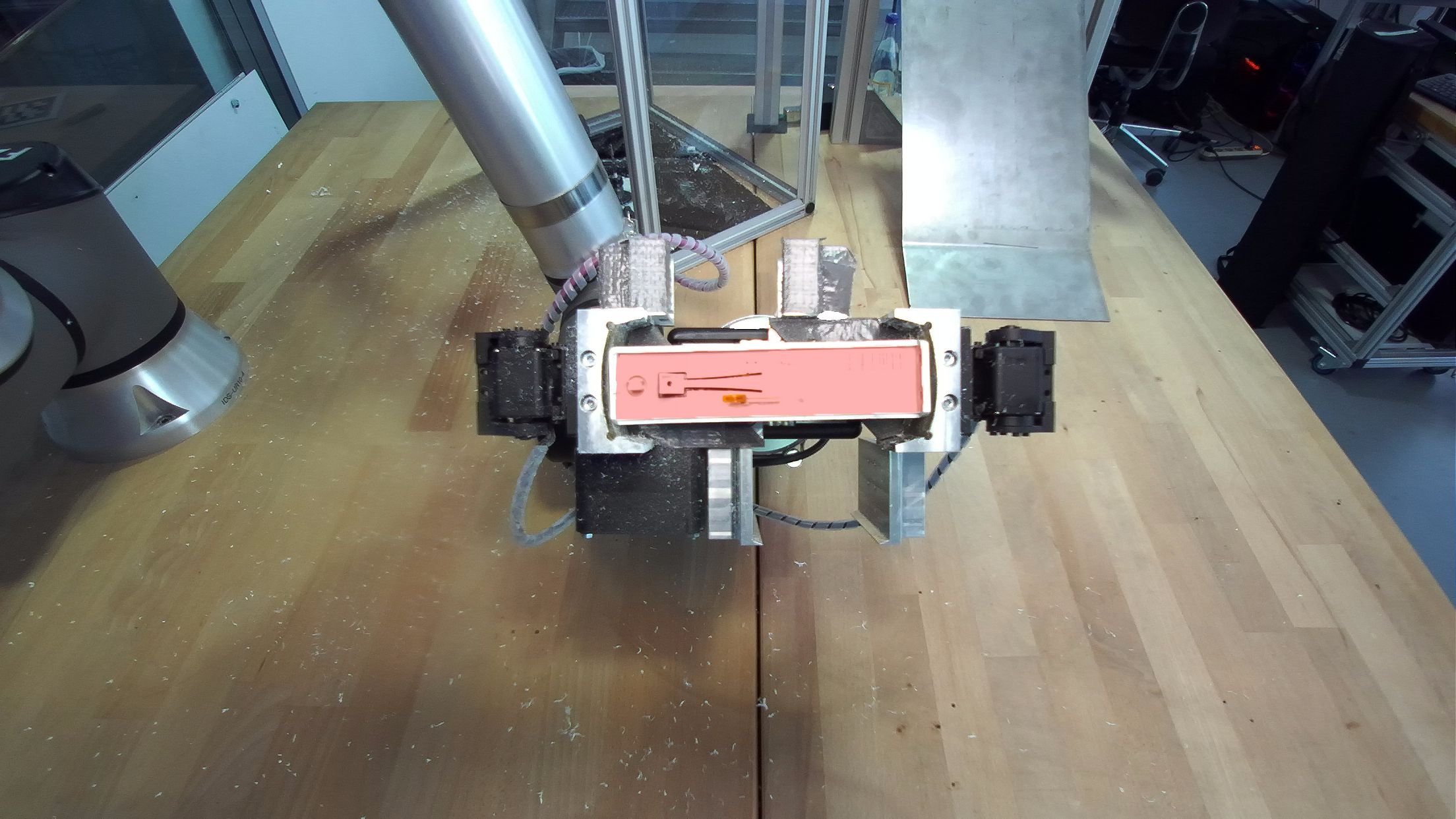}
    };
    \node[image] (E3) at ($(E2.north east) + (\imagespacing pt, 0)$) {
        \includegraphics[width=\imagewidth pt]{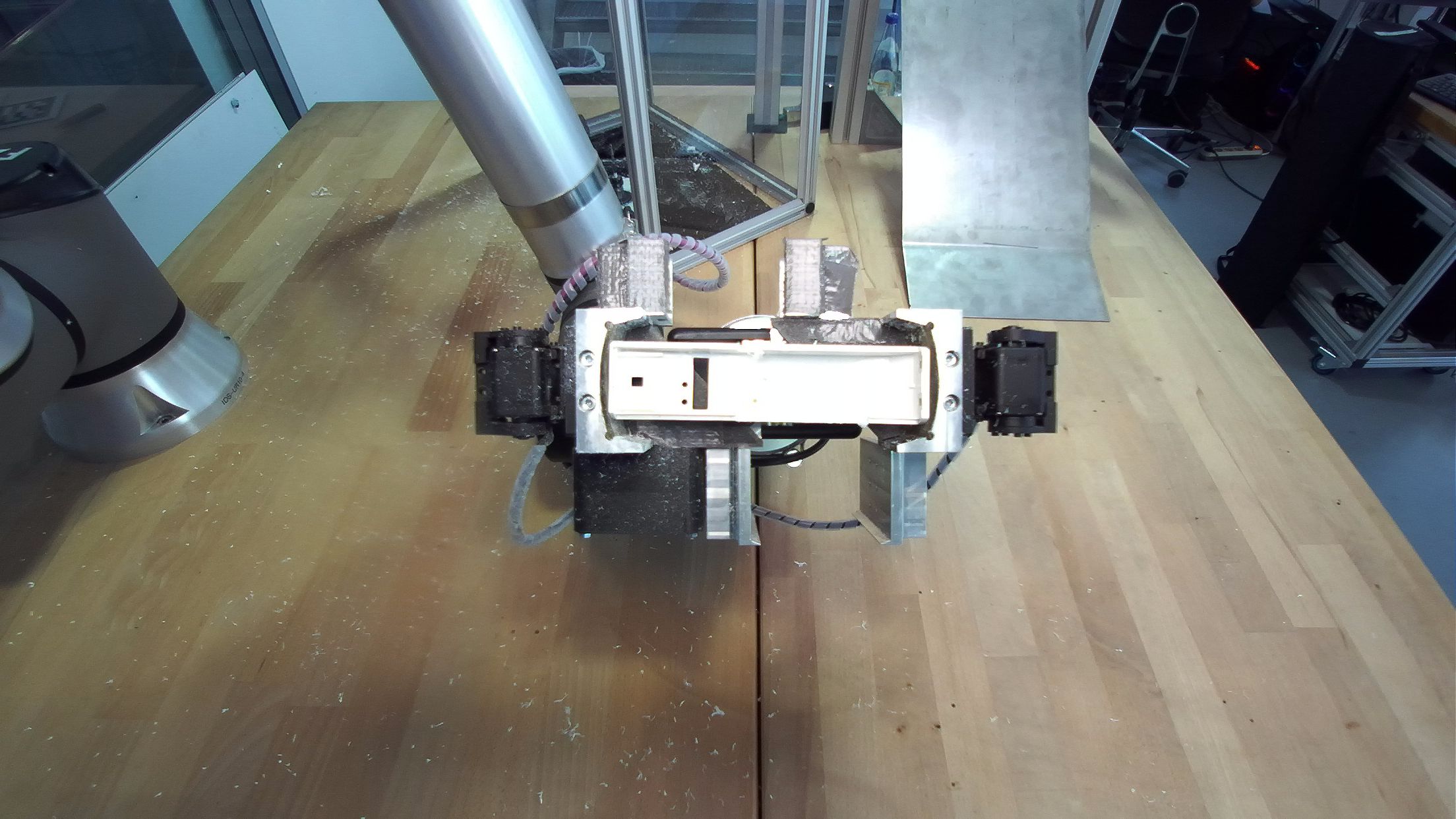}
    };
    \node[image] (E4) at ($(E3.north east) + (\imagespacing pt, 0)$) {
        \includegraphics[width=\imagewidth pt]{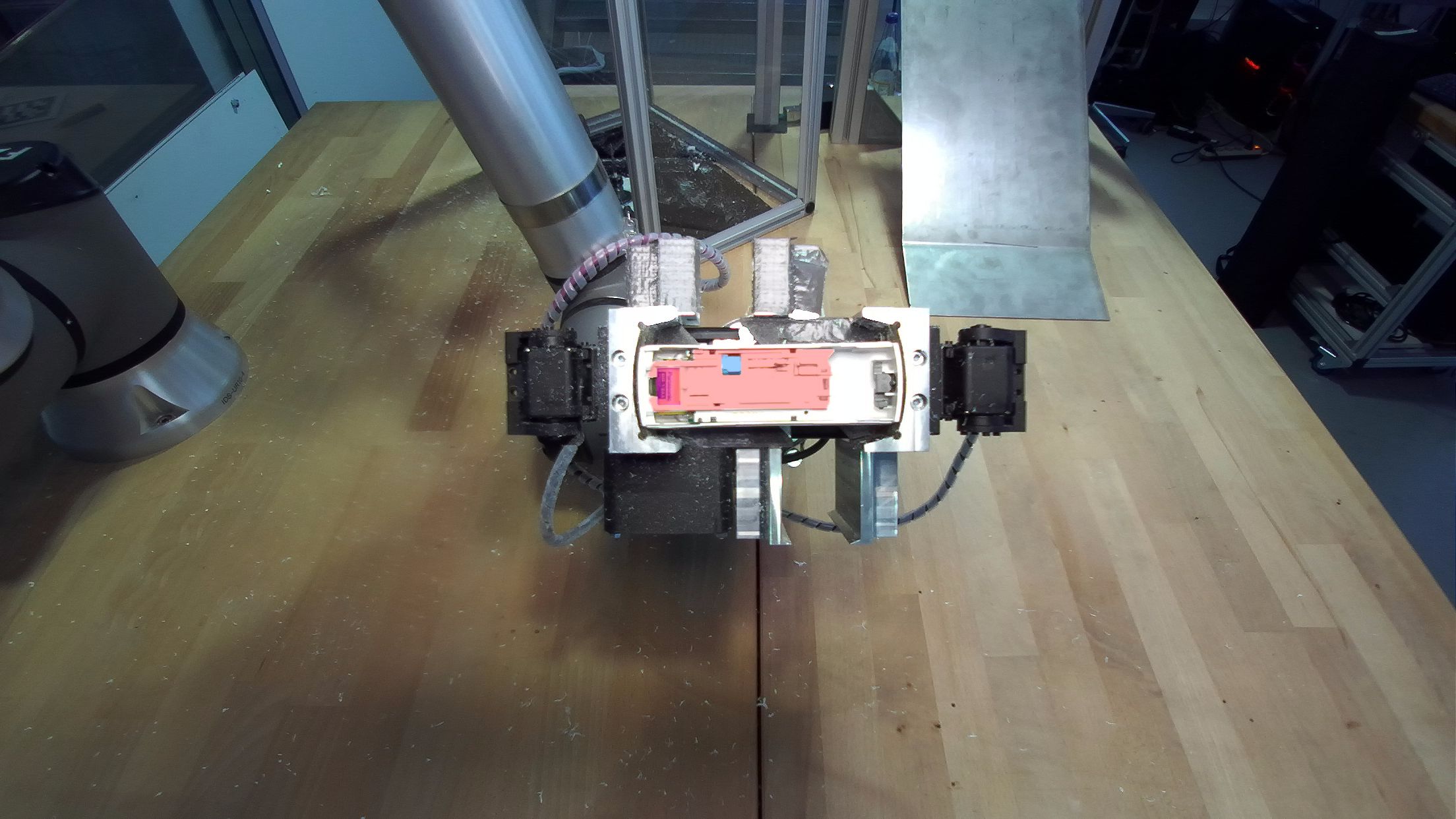}
    };
\end{tikzpicture}

%% file: figures/setup.tex
\begin{tikzpicture}[setup]
    \node[anchor=south west] (setup) at (0, 0) {
        \includegraphics[width=\linewidth]{
            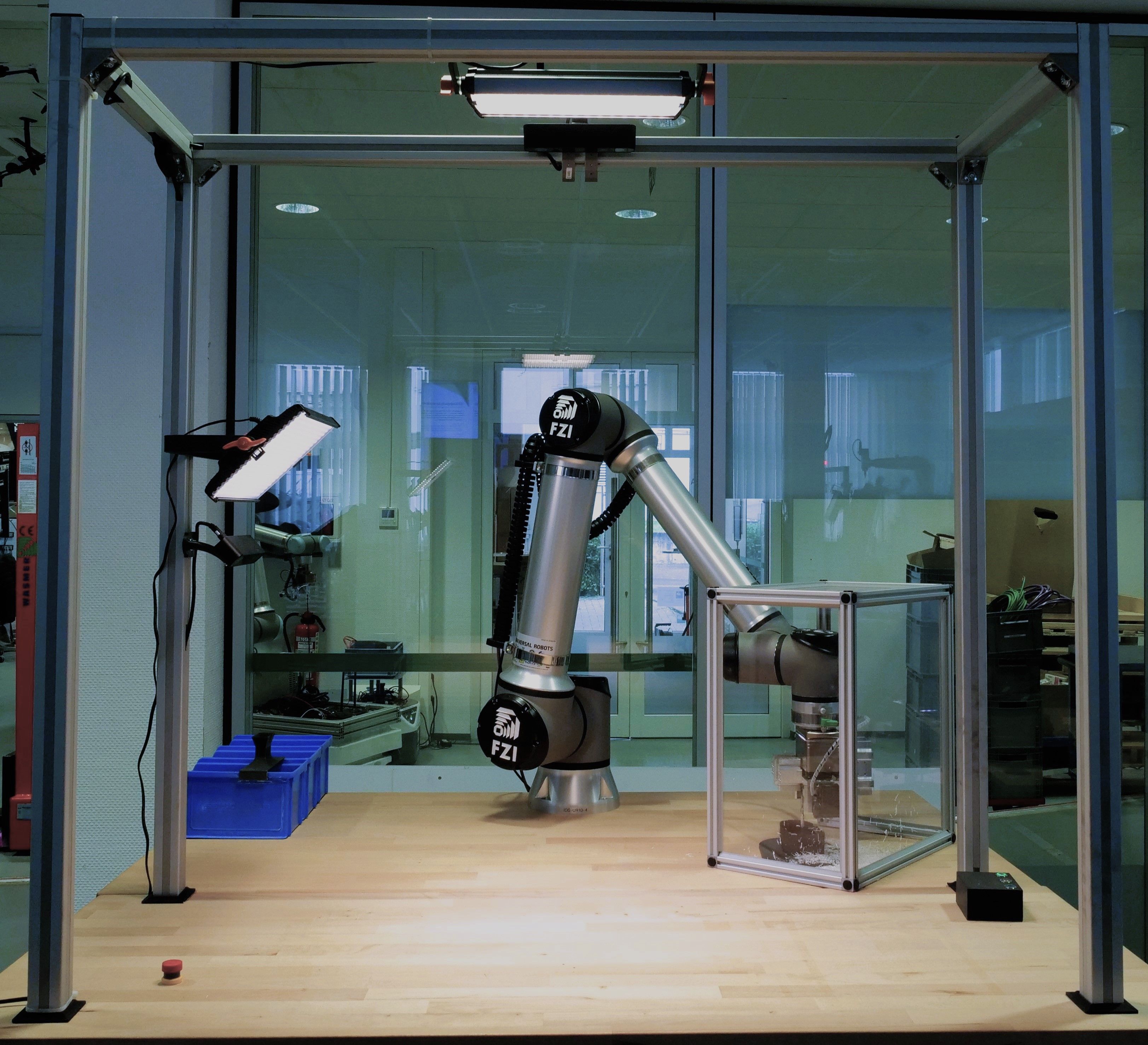
        }
    };

    \draw[color=red, very thick] (1.75, 3.75) circle (.5);
    \draw[color=red, very thick] (4.37, 6.8) circle (.5);
    \draw[color=green, very thick] (6, 1.65) circle (.5);
\end{tikzpicture}

%% file: sections/results.tex
\section{Results}
\label{sec:result}
An overview of the results of the ablation study is shown in Table~\ref{tab:results}, using the metrics defined in the previous section.
$D$ indicates the addition of natural language descriptions, $V$ the inclusion of the Vision Preprocessing Module, and $S$ the use of the \textit{dynamic preprocessing selection}.
The table reports results for each of the five tasks and their combined average.

The Vision Preprocessing Module produces a marked increase in Planning Accuracy across all tasks.
This is primarily attributable to the improved grounding of the objects in the semantics used in the task instructions, without which the VLM would often use incorrect location labels to attempt to interact with the objects.
As the \textit{dynamic preprocessing selection} $(S)$ is only used for verification, it does not affect the planning accuracy.

For Verification Accuracy, the baseline performance is approximately $50.0\%$, as the individual verification decisions are binary.
As with planning, the Vision Preprocessing Module $(V)$ greatly improves the verification accuracy.
As predicted, the \textit{dynamic preprocessing selection} $(S)$ further improves the verification accuracy for particularly object-dense environments, while not increasing the verification duration significantly.
In fact, in most cases, the VLM only requested a single preprocessed image containing exactly the objects mentioned in the goal.

At the end-to-end level, the complete pipeline achieves an Overall Accuracy of $67.3\%$.
However, we note that none of these planning errors concerned the semantic grounding of the objects but rather issues with instruction-following (such as skipping part of the task) or logical correctness (such as not opening the gripper before attempting to grasp an object).
The primary failure mode therefore shifted from semantic grounding to high-level task reasoning.

The evaluation of task \textbf{C} is of particular interest, as it requires performing facial recognition to differentiate between customers.
The baseline achieves no successful plans this task.
However, once the recognition is provided by the vision module, the task is solved with perfect planning accuracy.
Natural language descriptions initially also improve the planning accuracy (by $36.4\%$), such as for distinguishing between a blonde and a brunette.
However, more minute details are difficult to convey using language alone.

Also noteworthy is the performance on task \textbf{D}, which involves a large number of interactable objects in the cockpit setting.
The smaller improvement obtained from simultaneously masking all cockpit controls is consistent with the VLM being distracted by the large number of visual annotations.
The \textit{dynamic preprocessing selection} $(S)$ increases verification accuracy by $19.5$ percentage points by presenting masks only for objects relevant to the current goal.

\begin{table}
    \centering
    \vspace{1.2ex}
    \caption{Overview of the results of the ablation study, by task.}
    \input{figures/results}
    \label{tab:results}
\end{table}

\subsection*{Cross-Model Evaluation}
As shown in Table~\ref{tab:transfer}, the results obtained with the Gemma~3 VLM are generally consistent with those obtained with LLaVA-OneVision, albeit slightly higher.
Nonetheless, the same trends can be observed, with the Vision Preprocessing Module $(V)$ providing a substantial boost in both planning and verification accuracies, and the \textit{dynamic preprocessing selection} $(S)$ further improving the verification accuracy.
This experiment provides preliminary evidence that the benefits of the grounding interface are not specific to the primary VLM.

\subsection*{Discussion}
\label{sec:limitations}
The use of vision modules for preprocessing the VLM's visual input shifts the generalization requirements of the pipeline from the VLM to the dedicated vision model.
While this may seem like a minor change, it is important to note that the vision modules can be trained on much smaller datasets than VLMs, and can be easily fine-tuned to new domains with only a few additional examples.
Therefore, this shift is a key strength of the pipeline, and significantly increases its modularity and its practical applicability in specialized domains.

The controlled image-sequence evaluation uses manually prepared perception outputs and therefore isolates their effect on downstream planning and verification rather than measuring sensitivity to perception errors.
The real-robot case study complements this evaluation by demonstrating one complete implementation using learned detection, segmentation, pose estimation, and physical execution.
A natural extension would be to expose perception confidence through the same modular interface, allowing uncertain detections to trigger additional observations or controlled recovery actions.

After visual grounding is provided, the remaining planning failures primarily concern instruction following and logical consistency.
This indicates that grounding alone cannot compensate for all limitations of the reasoning model.
These errors could be addressed without changing the perception interface, for example through explicit action constraints, validation of generated Behavior Trees, or a structured memory for longer tasks.
Within its current scope, \shot~is best suited to structured applications in which domain-specific perception and robot skills are available and the environment changes primarily between actions.
For such applications, the proposed architecture provides a practical, flexible, and robust approach to combine evolving VLM capabilities with mature robot-control software and targeted domain knowledge.

\begin{table}
    \centering
    \vspace{1.2ex}
    \caption{Cross-model evaluation on task \textbf{A} using Gemma~3.}
    \input{figures/transfer}
    \label{tab:transfer}
\end{table}

%% file: figures/results.tex
\newcommand{\ditto}{\texttt{-\raisebox{-.4ex}{"}-}}
\newcommand{\taskrow}[1]{\multirow{5}{5em}{\centering\textbf{#1}}}

\begin{tabular}{l l ccc}
    \hline
                                & \multirow{2}{6em}{\textbf{Components}}    & \multicolumn{3}{c}{\textbf{Accuracy} [\%]}            \\
                                &                                           & \textbf{Plan.} & \textbf{Verif.} & \textbf{Overall}   \\
    \hline
    \taskrow{A\\Workshop}       & Baseline                                  & $0.0$          & $50.9$          & $0.0$              \\
                                & Descriptions $D$                          & $0.0$          & $52.8$          & $0.0$              \\
                                & Vision $V$                                & $72.7$         & $81.1$          & $27.3$             \\
                                & $D + V$                                   & $72.7$         & $81.1$          & $27.3$             \\
                                & $D + V + $ Selection $S$                  & \ditto         & $90.6$          & $54.6$             \\
    \hline
    \taskrow{B\\Agriculture}    & Baseline                                  & $0.0$          & $52.0$          & $0.0$              \\
                                & $D$                                       & $27.3$         & $60.0$          & $0.0$              \\
                                & $V$                                       & $90.9$         & $84.0$          & $45.5$             \\
                                & $D + V$                                   & $81.8$         & $86.0$          & $54.6$             \\
                                & $D + V + S$                               & \ditto         & $94.0$          & $81.8$             \\
    \hline
    \taskrow{C\\Service}        & Baseline                                  & $0.0$          & $47.2$          & $0.0$              \\
                                & $D$                                       & $36.4$         & $55.6$          & $9.1$              \\
                                & $V$                                       & $100$          & $94.4$          & $81.8$             \\
                                & $D + V$                                   & $100$          & $94.4$          & $81.8$             \\
                                & $D + V + S$                               & \ditto         & $94.4$          & $81.8$             \\
    \hline
    \taskrow{D\\Aerospace}      & Baseline                                  & $0.0$          & $51.2$          & $0.0$              \\
                                & $D$                                       & $0.0$          & $48.8$          & $0.0$              \\
                                & $V$                                       & $54.6$         & $61.0$          & $9.1$              \\
                                & $D + V$                                   & $54.6$         & $61.0$          & $18.2$             \\
                                & $D + V + S$                               & \ditto         & $80.5$          & $27.3$             \\
    \hline
    \taskrow{E\\Recycling}      & Baseline                                  & $18.2$         & $48.7$          & $0.0$              \\
                                & $D$                                       & $36.4$         & $54.1$          & $0.0$              \\
                                & $V$                                       & $100$          & $86.5$          & $54.6$             \\
                                & $D + V$                                   & $100$          & $89.2$          & $63.6$             \\
                                & $D + V + S$                               & \ditto         & $97.3$          & $90.9$             \\
    \hline
    \taskrow{Combined}          & Baseline                                  & $3.6$          & $50.2$          & $0.0$              \\
                                & $D$                                       & $20.0$         & $54.4$          & $1.8$              \\
                                & $V$                                       & $83.6$         & $81.1$          & $43.6$             \\
                                & $D + V$                                   & $81.8$         & $82.0$          & $49.1$             \\
                                & $D + V + S$                               & \ditto         & $91.2$          & $67.3$             \\
    \hline
    
\end{tabular}

%% file: figures/transfer.tex
\newcommand{\ditto}{\texttt{-\raisebox{-.4ex}{"}-}}
\newcommand{\taskrow}[1]{\multirow{5}{5em}{\centering\textbf{#1}}}

\begin{tabular}{l l ccc}
    \hline
                                    & \multirow{2}{6em}{\textbf{Components}}    & \multicolumn{3}{c}{\textbf{Accuracy} [\%]}            \\
                                    &                                           & \textbf{Plan.} & \textbf{Verif.} & \textbf{Overall}   \\
    \hline
    \taskrow{A\\Workshop\\(Gemma 3)}& Baseline                                  & $18.2$         & $60.4$          & $0.0$              \\
                                    & Descriptions $D$                          & $27.3$         & $66.0$          & $0.0$              \\
                                    & Vision $V$                                & $81.8$         & $81.1$          & $45.5$             \\
                                    & $D + V$                                   & $81.8$         & $84.9$          & $45.5$             \\
                                    & $D + V + $ Selection $S$                  & \ditto         & $94.3$          & $63.6$             \\
    \hline
    
\end{tabular}

%% file: sections/conclusion.tex
\section{Conclusion}
\label{sec:conclusion}
We presented \shot, a modular system that connects general-purpose VLM reasoning with domain-specific perception and established robot execution components.
The system incrementally generates and verifies short-horizon plans, while a Vision Preprocessing Module supplies semantic object labels, visual masks, and executable target poses.
For object-dense scenes, dynamic preprocessing selection allows the VLM to request only the information relevant to the current verification step.

Across five specialized tasks, the complete system achieves $81.8\%$ Planning Accuracy, $91.2\%$ Verification Accuracy, and $67.3\%$ Overall Accuracy, compared with $3.6\%$, $50.2\%$, and $0.0\%$ for the ungrounded baseline.
The brownfield industrial e-waste disassembly application further demonstrates how the architecture can reuse ROS~2, MoveIt~2, Behavior Trees, and targeted perception models without task-specific VLM fine-tuning.
The results show that explicit, modular visual grounding provides a practical route for applying VLM-based task planning in specialized robotic applications while retaining existing robot software and domain-specific capabilities.

%% file: sections/acknowledgments.tex
\section{Acknowledgments}
    We would like to thank our colleagues at \xblackout{the FZI Research Center for Information Technology} for their valuable feedback and support.
    Special thanks to \xblackout{Vincenzo Di Pentima}, for \ifdefined\review their \else his \fi tireless work on the robot hardware and software setup, 
    as well as to \xblackout{Martin Schulze}, for helpful discussions on the current state of the art, 
    and \xblackout{David Oberacker} and \xblackout{Niklas Spielbauer}, for their insightful early reviews of this paper.